\documentclass[final,1p,times,numbers,sort&compress]{elsarticle}

\usepackage{graphicx}
\usepackage{mathbbold}
\usepackage{subfigure}
\usepackage{amsmath}
\usepackage{amssymb}
\usepackage{multirow}
\usepackage{chemsym}
\usepackage{algorithm}
\usepackage{algorithmic}

\newtheorem{theorem}{Theorem}[section]
\newtheorem{proposition}[theorem]{Proposition}
\newtheorem{definition}[theorem]{Definition}
\newtheorem{lemma}[theorem]{Lemma}

\newproof{pf}{Proof}

\journal{}

\begin{document}

\begin{frontmatter}



\title{Fixed Point Algorithm Based on Quasi-Newton Method for Convex Minimization Problem with Application to Image Deblurring}


\author[dqc,nudt]{Dai-Qiang Chen et.al\corref{cor1}}
\cortext[cor1]{Corresponding author}
\ead{chener050@sina.com}


\address[dqc]{Department of Mathematics, School of Biomedical Engineering, Third Military Medical University, Chongqing 400038, Chongqing, People's Republic of China}

\address[nudt]{Department of Mathematics and System, School of Sciences, National University of Defense Technology, Changsha 410073, Hunan, People's Republic of China}

\begin{abstract}
Solving an optimization problem whose objective function is the sum of two convex functions has received
considerable interests in the context of image processing recently. In particular, we are interested in the scenario when a non-differentiable convex function such as the total variation (TV) norm is included in the objective function due to many variational models established in image processing have this nature. In this paper, we propose a fast fixed point algorithm based on the quasi-Newton method for solving this class of problem, and apply it in the field of TV-based image deblurring. The novel method is derived from the idea of the quasi-Newton method, and the fixed-point algorithms based on the proximity operator, which were widely investigated very recently. Utilizing the non-expansion property of the proximity operator we further investigate the global convergence of the proposed algorithm. Numerical experiments on image deblurring problem with additive or multiplicative noise are presented to demonstrate that the proposed algorithm is superior to the recently developed fixed-point algorithm in the computational efficiency.

\end{abstract}

\begin{keyword}


Newton method; primal-dual; fixed-point algorithm; total variation; Rayleigh noise
\end{keyword}

\end{frontmatter}



\section{Introduction}\label{sec1}

The general convex optimization problems that arise in image processing take the form of a sum of two convex functions. Often one function is the data fidelity energy term that is decided by the noise type and one wants to minimize, and the other function is the regularization term to make the solution have certain properties. For instance, the usual $l_{1}$ based regularization is used to obtain the sparse solution in the fields such as image restoration \cite{CVPR:TVwavelet} and compressed sensing \cite{IEEEIT:CandesCS, IEEEIT:DonohoCS}. In this paper, we propose an efficient fixed-point algorithm to solve the optimization problem whose objective function is composed of two convex function, i.e.,
\begin{equation}\label{equ1.1}
\min\limits_{u} f_{1}(B u) + f_{2}(u)
\end{equation}
where $f_{1}: \mathbb{R}^{M}\rightarrow \mathbb{R}$ and $f_{2}: \mathbb{R}^{N}\rightarrow \mathbb{R}$ are convex function, $B: \mathbb{R}^{N}\rightarrow \mathbb{R}^{M}$ is a linear transform, and $f_{2}$ is differentiable with a $1/\beta$-Lipschitz continuous gradient, i.e.,

\begin{equation}\label{equ1.6}
\langle \nabla f_{2}(v)-\nabla f_{2}(w), v-w\rangle\geq \beta\|\nabla f_{2}(v)-\nabla f_{2}(w)\|^{2}_{2}
\end{equation}
for any $v, w\in \mathbb{R}^{N}$. Despite its simplicity, many variational models in image processing can be formulated in the form of (\ref{equ1.1}). For example, the classical total variation (TV) or wavelet sparsity prior based models \cite{CVPR:TVwavelet}, which are often considered in image restoration under Gaussian noise, have the simple form as follows.
\begin{equation}\label{equ1.2}
\min\limits_{u} \mu \|B u\|_{1} + \frac{1}{2}\|K u - b\|_{2}^{2}
\end{equation}
where $K\in \mathbb{R}^{N\times N}$ is a linear blurring operator, $b\in \mathbb{R}^{N}$ and $\mu>0$ is a regularization parameter. Here $B$ denotes the sparse transform such as the gradient operator and the wavelet basis, and the image prior is imposed by using the term $\|B u\|_{1}$, which promotes the sparsity of image under the transform $B$. With $f_{1}=\mu \|\cdot\|_{1}$ and $f_{2}=\frac{1}{2}\|K \cdot - b\|_{2}^{2}$ the problem (\ref{equ1.2}) can be seen as a special case of (\ref{equ1.1}). It is observed that the main difficulty of solving problem (\ref{equ1.1}) is that the function $f_{1}$ is non-differentiable.

In the last several years, many optimization algorithms have been developed for efficiently solving the variational models in image processing. The iterative shrinkage/thresholding (IST) algorithm is one of the most successful methods. Consider the general minimization problem

\begin{equation}\label{equ1.3}
\min\limits_{u} f(u) + g(u)
\end{equation}
where $f$ and $g$ are convex function, and $g$ is differentiable. The classical IST algorithm for problem (\ref{equ1.3}) is given by the following iterative formula

\begin{equation}\label{equ1.4}
u^{k+1}=\mathcal{T}_{\gamma f}(u^{k}-\gamma \nabla g(u^{k}))=p_{\gamma}(u^{k})
\end{equation}
where $\gamma>0$ is a step parameter. Here $\mathcal{T}_{\gamma f}$ is called the thresholding operator. It corresponds to the proximity operator $\textrm{prox}_{\gamma f}$, which is defined by \cite{ProxDefine}

\begin{equation}\label{equ1.5}
\textrm{prox}_{\gamma f}(v)=\textrm{arg}\min\limits_{u} \gamma f(u) + \frac{1}{2}\|u-v\|_{2}^{2}.
\end{equation}
In different literatures, IST is also called iterative denoising method \cite{IEEETIP:EMwavelet}, Landweber iteration \cite{CPAM:IST}, proximal forward-back splitting (PFBS) algorithm \cite{SIAM:PFBS} or fixed-point continuation (FPC) algorithm \cite{RiceReport:FPC}. In order to further accelerate the convergence speed, many new iterative shrinkage algorithms based on the IST, which include the SpaRSA (Sparse Reconstruction by Separable Approximation) \cite{IEEETSP:SpaRSA}, TwIST (Two-step IST) \cite{IEEETIP:TwIST}, FISTA (Fast iterative shrinkage-thresholding algorithm) \cite{SIAM:FISTA} were further proposed. Notice that a proximity operator is needed to be computed in each iteration of the iterative shrinkage algorithms. However, the proximity operators $\textrm{prox}_{\gamma f}$ for the general case of $f = f_{1}\circ B$ often have no closed solutions. For example, if we choose $B=\nabla$ and $f = \|\nabla \cdot\|_{1}$, then the minimization problem of (\ref{equ1.5}) is just the Rudin-Osher-Fatemi (ROF) denoising problem whose solution cannot be obtained easily. Therefore, inner iterative algorithm is needed for computing the proximity operators in most cases.

In recent years, a class of algorithms based on the splitting methods have been developed and shown to be efficient for computing the proximity operator. For instance, Goldstein and Osher \cite{SIAM:SplitBregman} proposed a splitting algorithm based on the Bregman iteration, called the split-Bregman method, to compute the solution of the minimization problem of (\ref{equ1.5}) especially for the case of ROF denoising. This algorithm can be successfully applied for solving the general minimization problem (\ref{equ1.1}), and theoretically it has been proved to be equivalent to the Douglas-Rachford splitting (DRS) algorithm \cite{IEEESTSP:DRS, IJCV:DRS} and the alternating direction of multiplier method (ADMM) \cite{UCLA:ADMM, JSC:TVStokes}. Although the split-Bregman framework has been shown to be very useful, a sub-minimization problem of solving the system of linear or nonlinear equations is included in each iteration and may time-consuming sometimes. Very recently, alternating direction minimization methods based on the linearized technique \cite{JSC:LADM, SIAMImage:GILADM} have been widely investigated to overcome this and further improve the efficiency. Another class of methods is the primal-dual algorithms. Chambolle \cite{JMIV:ChambollePD} firstly proposed a dual algorithm for the ROF denoising. Later on, Zhu et al. \cite{UCLAReport:PDHG} devised a primal-dual hybrid gradient (PDHG) method, which alternately update the primal and dual variables by the gradient descent scheme and gradient ascend scheme. The theoretical analysis on variants of the PDHG algorithm, and on the connection with the linearized version or variants of ADMM were widely investigated to bridge the gap between different methods. Refer to \cite{JSC:LADM, JMIV:PDHG, ICCV:DPDHG, SIAM:GFPDHG, SIAM:PDHGSP} and the references cited therein for details.

In this paper, we focus our attention on a new class of algorithms that has been developed very recently from the view of fixed-point. In \cite{ACM:FastAlg}, Jia and Zhao proposed a fast algorithm for the ROF denoising by simplify the original split-Bregman framework. Motivated by this idea, Micchelli et al. \cite{IP:FP2O} designed a fixed-point algorithm based on proximity operator (named $\textrm{FP^{2}O}$) for computing $\textrm{prox}_{f_{1}\circ B}$, which was proved to be more efficient than the splitting methods. Later on, several variants of fixed-point algorithms were proposed for special cases of image restoration. For instance, Micchelli et al. \cite{ACM:FP2OTVL1, IPI:FP2OTVL1} further extended the $\textrm{FP^{2}O}$ algorithm to solve $\textrm{TV-L1}$ denoising model where $f_{2}=\frac{1}{2}\|K \cdot - b\|_{1}$ in (\ref{equ1.1}). Chen et al. \cite{UCLAReport:PDHG} proposed a proximity operator based algorithm for solving indicator functions based $l_{1}$-norm minimization problems with application to compressed sample. Krol et al. \cite{IP:PAPA} proposed a preconditioned alternating projection algorithm for emission computed tomography (ECT) restoration, where a diagonal preconditioning matrix is used in the devised fixed-point algorithm. The extension of the $\textrm{FP^{2}O}$ algorithm to the more general case of the form of (\ref{equ1.1}) has also been investigated very recently \cite{Arxiv:IFP2O, JMIV:IFP2O}. Specifically, a primal-dual fixed point algorithm which combines the PFBS algorithm and only one inner iteration of $\textrm{FP^{2}O}$ has been proposed in \cite{IP:PDFP2O}.

However, in the previous fixed-point algorithms, we observe that all the iterative formulas are composed of the gradient descent algorithm and the proximity point algorithm. It is well known that the gradient-based algorithms typically have a sub-linear convergence rate, while the Newton method or the quasi-Newton method has been presented with a super-linear convergence rate. This fact motivates us to propose a new fixed-point algorithm which combines the quasi-Newton method and the proximity operator algorithm. Furthermore, the global convergence of the proposed algorithm is investigated under certain assumption.

The rest of this paper is organized in four sections. In section \ref{sec2} we briefly review the existing fixed-point algorithms based on the proximal operator, and further propose a fixed-point algorithm based on quasi-Newton method. In section \ref{sec3} the global convergence of the proposed algorithm is further investigated from the point of the view of fixed point theory under certain conditions. The numerical examples on deblurring problem of images contaminated by additive Gaussian noise and multiplicative noise are reported in section \ref{sec4}. The results there demonstrate that $\textrm{FP^{2}O_{\kappa}\_QN}$ is superior to the recently proposed $\textrm{PDFP^{2}O_{\kappa}}$ in the context of image deblurring.

\section{Fixed-point algorithm based on quasi-Newton method}\label{sec2}
\setcounter{equation}{0}

\subsection{Existing fixed-point algorithms based on the proximal operator}\label{subsec2.1}

Motivated by the fast algorithm proposed for the ROF denoising in the literature \cite{ACM:FastAlg}, Micchelli et al. \cite{IP:FP2O} designed a fixed-point algorithm named $\textrm{FP^{2}O}$ for the computation of the proximity operator $\textrm{prox}_{f_{1}\circ B}(x)$ for any $x\in \mathbb{R}^{N}$. Denote $\lambda_{\max}(BB^{T})$ be the largest eigenvalue of $BB^{T}$. Choose the parameter $0<\lambda<2/\lambda_{\max}$, and define the operator

\begin{equation}\label{equ2.1}
S(v)=\left(I-\textrm{prox}_{\frac{1}{\lambda}f_{1}}\right)\left(B x+\left(I-\lambda BB^{T}\right)v\right).
\end{equation}
Then we can obtain the fixed-point iterative scheme which is just called $\textrm{FP^{2}O}$ algorithm as follows

\begin{equation}\label{equ2.2}
v^{k+1}=S_{\kappa}(v^{k})
\end{equation}
where $S_{\kappa}$ is the $\kappa$-averaged operator of $S$, i.e., $S_{\kappa}=\kappa I+(1-\kappa)S$ for any $\kappa\in (0, 1)$. Calculate the fixed-point $v^{*}$ of the operator $S$ by the formula (\ref{equ2.2}), and hence obtain that

\begin{equation}\label{equ2.3}
\textrm{prox}_{f_{1}\circ B}(x)=x-\lambda B^{T}v^{*}.
\end{equation}

The key technique served as the foundation of $\textrm{FP^{2}O}$ algorithm is the relationship between
the proximity operator and the subdifferential of a convex function, as described in (\ref{equ3.2}) below. $\textrm{FP^{2}O}$ algorithm supplies a simple and efficient method of solving (\ref{equ1.1}) with the special case of $f_{2}(u)=\frac{1}{2}\|u-x\|_{2}^{2}$ in the classical framework of fixed-point iteration. In \cite{JMIV:IFP2O}, this algorithm has been extended to the more general case that $\nabla f_{2}(u)$ is bijective and the inverse can be computed easily. In particular, choose $f_{2}(u)=\frac{1}{2}u^{T}Qu-x^{T}u$, where $Q$ is a positive definite $N\times N$ matrix. Then (\ref{equ1.1}) can be reformulated as

\begin{equation}\label{equ2.4}
\min\limits_{u} f_{1}(B u) + \frac{1}{2}u^{T}Qu-x^{T}u
\end{equation}
Define the operator

\begin{equation}\label{equ2.5}
\hat{S}(v)=\left(I-\textrm{prox}_{\frac{1}{\lambda}f_{1}}\right)\left(BQ^{-1} x+\left(I-\lambda BQ^{-1}B^{T}\right)v\right).
\end{equation}
Then the corresponding fixed-point iterative scheme for (\ref{equ2.4}) is given by
\begin{equation}\label{equ2.6}
v^{k+1}=\hat{S}_{\kappa}(v^{k}),
\end{equation}
and the solution $u^{*}$ of (\ref{equ2.4}) can be obtained by the formula

\begin{equation}\label{equ2.7}
u^{*}=Q^{-1}\left(x-\lambda B^{T}v^{*}\right)
\end{equation}
where $v^{*}$ is the fixed-point of the operator $\hat{S}_{\kappa}$.

In order to deal with the general case of $f_{2}$, the authors in \cite{Arxiv:IFP2O} also combined $\textrm{FP^{2}O}$ and PFBS algorithms, and proposed a new algorithm named $\textrm{PFBS\_FP^{2}O}$ in which the proximity operator in the PFBS algorithm is calculated by using $\textrm{FP^{2}O}$, i.e.,
\[
u^{k+1}=\textrm{prox}_{\gamma f_{1}\circ B}\left(u^{k}-\gamma \nabla f_{2}(u^{k})\right)
\]
is calculated by $\textrm{FP^{2}O}$. Notice that a inner iteration of solving $\textrm{prox}_{\gamma f_{1}\circ B}$ is included in $\textrm{PFBS\_FP^{2}O}$, and it is problematic to set the approximate iteration number to balance the computational time and precision. In order to solve this issue, a primal-dual fixed points algorithm based on proximity operator ($\textrm{PDFP^{2}O}$) \cite{IP:PDFP2O} was proposed very recently. In this algorithm, instead of implementing $\textrm{FP^{2}O}$ for many iteration steps to calculate $\textrm{prox}_{\gamma f_{1}\circ B}(x)$ in $\textrm{PFBS\_FP^{2}O}$, only one inner fixed-point iteration is adopted. Suppose $\kappa=0$. Then we can obtain the following iteration scheme ($\textrm{PDFP^{2}O}$):

\begin{equation}\label{equ2.8}
\left\{\begin{array}{lll}u^{k+1/2} = u^{k}-\gamma \nabla f_{2}(u^{k}), &~ &~ ~\\
v^{k+1} = \left(I-\textrm{prox}_{\frac{\gamma}{\lambda}f_{1}}\right)\left(B u^{k+1/2}+\left(I-\lambda BB^{T}\right)v^{k}\right),
&~ & ~ ~\\
u^{k+1} = u^{k+1/2}-\lambda B^{T}v^{k+1}.
\end{array} \right.
\end{equation}
It is obvious that $u$ is the primal variable related to (\ref{equ1.1}), and according to the thorough study in \cite{IP:PDFP2O} we know that the variable $v$ is just the dual variable of the primal-dual form related to (\ref{equ1.1}). Therefore, $\textrm{PDFP^{2}O}$ also belongs to the class of primal-dual algorithm framework. Similarly to $\textrm{FP^{2}O}$, a relaxation parameter $\kappa\in (0,1)$ can be introduced to get the algorithm named $\textrm{PDFP^{2}O_{\kappa}}$. For more details refer to \cite{IP:PDFP2O}.

\subsection{Proposed fixed-point algorithm based on quasi-Newton method}\label{subsec2.2}

In the $\textrm{PDFP^{2}O}$ algorithm, the iterative formulas consist of the proximity operator and the gradient descent algorithm. Since the Newton-type methods have been shown to have a faster convergence rate compared to the gradient-based methods, a very nature idea is to use the Newton-type methods instead of the gradient descent step in the fixed-point algorithm. Consider the minimization problem (\ref{equ1.1}). We use the second-order Taylor expansion of the convex function $f_{2}(u)$ at the recent iterative point $u^{k}$ instead of it, i.e.,
\[
f_{2}(u)\approx f_{2}(u^{k})+\langle \nabla f_{2}(u^{k}), u-u^{k}\rangle + \frac{1}{2}\left(u-u^{k}\right)^{T}Q^{k}\left(u-u^{k}\right)
\]
where $Q^{k}$ is a positive definite symmetric matrix to approximate the second derivative $\nabla^{2}f_{2}$.
Then (\ref{equ1.1}) can be reformulated as

\begin{equation}\label{equ2.9}
\min\limits_{u} f_{1}(B u) + \frac{1}{2}u^{T}Q^{k}u- (Q^{k}u^{k}-\nabla f_{2}(u^{k}))^{T}u.
\end{equation}
It is observed that (\ref{equ2.9}) corresponds to the minimization problem (\ref{equ2.4}) with $x=Q^{k}u^{k}-\nabla f_{2}(u^{k})$. Therefore, the next iteration scheme $u^{k+1}$ can be obtained by the fixed point iteration algorithm shown in (\ref{equ2.5})--(\ref{equ2.7}). In order to avoid any inner iterations, we use only one inner fixed point iteration in the proposed algorithm. For this, choose $\kappa=0$, and define
\[
S_{k+1}(v)=\left(I-\textrm{prox}_{\frac{1}{\lambda}f_{1}}\right)\left(B(u^{k}-(Q^{k})^{-1}\nabla f_{2}(u^{k}))+\left(I-\lambda B(Q^{k})^{-1}B^{T}\right)v\right).
\]
Using the numerical solution $v^{k}$ for $S_{k}$ as the initial value, and only implementing one iteration of solving the fixed-point of $S_{k+1}(v)$, we can obtain the following iteration scheme

\begin{equation}\label{equ2.10}
\left\{\begin{array}{lll}
v^{k+1} = \left(I-\textrm{prox}_{\frac{1}{\lambda}f_{1}}\right)\left(B(u^{k}-(Q^{k})^{-1}\nabla f_{2}(u^{k}))+\left(I-\lambda B(Q^{k})^{-1}B^{T}\right)v^{k}\right),
&~ & ~ ~\\
u^{k+1} = u^{k}-(Q^{k})^{-1}\nabla f_{2}(u^{k})-\lambda (Q^{k})^{-1}B^{T}v^{k+1}.
\end{array} \right.
\end{equation}
Setting $u^{k+1/2}=u^{k}-(Q^{k})^{-1}\nabla f_{2}(u^{k})$. It is observed that an intermediate iterative variable $u^{k+1/2}$ is generated by a quasi-Newton method. Therefore, we called the proposed algorithm a fixed-point algorithm based on quasi-Newton method, and abbreviate it as $\textrm{FP^{2}O\_QN}$, which is described as \textbf{Algorithm 1} below. For simplification of convergence analysis below, we set $Q^{k}\equiv Q$ to be unchanged with different $k$.

\begin{algorithm}[htb]
\caption{ Fixed-point algorithm based on quasi-Newton method ($\textrm{FP^{2}O\_QN}$)}
\begin{algorithmic}[1]
\REQUIRE    set $u_{0}\in \mathbb{R}^{N}$, $v_{0}\in \mathbb{R}^{M}$, $0< \lambda\leq 1/\lambda_{\max}(BQ^{-1}B^{T})$.\\
\textbf{Main Iteration}: \\
~~~$u^{k+1/2}=u^{k}-Q^{-1}\nabla f_{2}(u^{k})$; \\
~~~$v^{k+1} = \left(I-\textrm{prox}_{\frac{1}{\lambda}f_{1}}\right)\left(Bu^{k+1/2}+\left(I-\lambda BQ^{-1}B^{T}\right)v^{k}\right)$; \\
~~~$u^{k+1} = u^{k+1/2}-\lambda Q^{-1}B^{T}v^{k+1}$. \\
\end{algorithmic}
\end{algorithm}

Similarly to the literatures \cite{IP:FP2O, JMIV:IFP2O, IP:PDFP2O}, we can introduce a relaxation parameter $\kappa\in (0, 1]$ to obtain \textbf{Algorithm 2}, which is exactly the Picard iterates with the parameter.

\begin{algorithm}[htb]
\caption{ $\textrm{FP^{2}O_{\kappa}\_QN}$}
\begin{algorithmic}[2]
\REQUIRE    set $u_{0}\in \mathbb{R}^{N}$, $v_{0}\in \mathbb{R}^{M}$, $0< \lambda\leq 1/\lambda_{\max}(BQ^{-1}B^{T})$.\\
\textbf{Main Iteration}: \\
~~~$u^{k+1/2}=u^{k}-Q^{-1}\nabla f_{2}(u^{k})$; \\
~~~$\hat{v}^{k+1} = S_{k+1}(v^{k})$; \\
~~~$\hat{u}^{k+1} = u^{k+1/2}-\lambda Q^{-1}B^{T}v^{k+1}$. \\
~~~$v^{k+1} = \kappa v^{k}+(1-\kappa)\hat{v}^{k+1}$; \\
~~~$u^{k+1} = \kappa u^{k}+(1-\kappa)\hat{u}^{k+1}$. \\
\end{algorithmic}
\end{algorithm}

\section{Convergence analysis}\label{sec3}
\setcounter{equation}{0}

Let us start with some related notations and conclusions which will serve as the foundation for the proof below.

\begin{definition}\label{def3.1}
(Nonexpansive operator) A nonlinear operator $T: \mathbb{R}^{M}\rightarrow \mathbb{R}^{M}$ is called nonexpansive if for any $x, y \in \mathbb{R}^{M}$,
\[
\|T(x)-T(y)\|_{2}\leq \|x-y\|_{2}.
\]
A nonlinear operator $P: \mathbb{R}^{M}\rightarrow \mathbb{R}^{M}$ is called firmly nonexpansive if for any $x, y \in \mathbb{R}^{M}$,
\[
\|P(x)-P(y)\|_{2}^{2}\leq \langle x-y, Px-Py \rangle.
\]
\end{definition}
By the application of the Cauchy-Schwarz inequality it is easy to show that a firmly nonexpansive operator is
also nonexpansive.

\begin{definition}\label{def3.2}
(Picard sequence \cite{BAM:Opial})For a given initial point $v^{0}\in \mathbb{R}^{M}$ and an operator $P: \mathbb{R}^{M}\rightarrow \mathbb{R}^{M}$, the sequence $\{v^{k}: k\in \mathbb{N}\}$ generated by $v^{k+1}=P(v^{k})$ is called the Picard sequence of the operator $P$.
\end{definition}
For the Picard sequence we have the following conclusion.

\begin{proposition}\label{pro2}
(Opial $\kappa$-averaged Theorem \cite{BAM:Opial}) Let $C$ be a closed convex set in $\mathbb{R}^{M}$ and let $P: C\rightarrow C$ be a nonexpansive mapping with at least one fixed point. Then for any $v^{0}\in R^{M}$
and any $\kappa\in (0,1)$, the Picard sequence of $P_{\kappa}=\kappa I+(1-\kappa)P$ converges to a fixed point of $P$.
\end{proposition}

For any convex function $f: \mathbb{R}^{N}\rightarrow \mathbb{R}\bigcup \{+\infty\}$, the
subdifferential of $f$ at $x\in \mathbb{R}^{N}$ is defined by

\begin{equation}\label{equ3.1}
\partial f(x)=\left\{y\in \mathbb{R}^{N}: f(z)\geq f(x)+ \langle y,z-x \rangle, ~~\forall~ z\in
\mathbb{R}^{N}\right\}.
\end{equation}
The following result illustrates the relationship between the proximity operator and the subdifferential
of a convex function. This conclusion has appeared in many previous literatures, such as \cite{IP:FP2O, JMIV:IFP2O, IP:PDFP2O}.

\begin{proposition}\label{pro1}
If $f$ is a convex function defined on $\mathbb{R}^{N}$ and $x, y\in \mathbb{R}^{N}$, then
\begin{equation}\label{equ3.2}
y\in \partial f(x) \Leftrightarrow x=\textrm{prox}_{f}(x+y).
\end{equation}
\end{proposition}

In what follows, we establish a fixed-point formulation for the solution of the minimization problem (\ref{equ1.1}) based on the conclusion in Proposition \ref{pro1}. To this end, we define the operator $T_{1}: \mathbb{R}^{M}\times \mathbb{R}^{N}\rightarrow \mathbb{R}^{M}$ as

\begin{equation}\label{equ3.3}
T_{1}(v,u)=\left(I-\textrm{prox}_{\frac{1}{\lambda}f_{1}}\right)\left(B(u-Q^{-1}\nabla f_{2}(u))+\left(I-\lambda BQ^{-1}B^{T}\right)v\right),
\end{equation}
and the operator $T_{2}: \mathbb{R}^{M}\times \mathbb{R}^{N}\rightarrow \mathbb{R}^{N}$ as

\begin{equation}\label{equ3.4}
T_{2}(v,u)=u-Q^{-1}\nabla f_{2}(u)-\lambda Q^{-1}B^{T}T_{1}
\end{equation}
where $\lambda$ is a positive parameter. Denote the operator $T: \mathbb{R}^{M}\times \mathbb{R}^{N}\rightarrow \mathbb{R}^{M}\times \mathbb{R}^{N}$ as
\begin{equation}\label{equ3.5}
T(v,u)=\left(T_{1}(v,u), T_{2}(v,u)\right).
\end{equation}

\begin{theorem}\label{the1}
If $u^{*}$ is a solution of the minimization problem (\ref{equ1.1}), then there exists $v^{*}\in R^{M}$ that satisfies:
\[
v^{*}=T_{1}(v^{*},u^{*}), ~~~~~u^{*}=T_{2}(v^{*},u^{*})
\]
which implies that $(v^{*},u^{*})$ is a fixed point of $T$. Conversely, if $w^{*}=(v^{*},u^{*})$ is a fixed point of $T$, then $u^{*}$ is a solution of the minimization problem (\ref{equ1.1}).
\end{theorem}

\begin{pf}
Since $u^{*}$ is one solution of the minimization problem (\ref{equ1.1}), by the first-order optimality condition we have
\[
\begin{split}
0\in \nabla f_{2}(u^{*})+(B^{T}\circ \partial f_{1} \circ B)(u^{*})\\
\Leftrightarrow 0\in Q^{-1}\nabla f_{2}(u^{*})+(Q^{-1}B^{T}\circ \partial f_{1} \circ B)(u^{*}) \\
\Leftrightarrow u^{*}\in u^{*}-Q^{-1}\nabla f_{2}(u^{*})-\lambda\left(Q^{-1}B^{T}\circ \frac{1}{\lambda}\partial f_{1} \circ B\right)(u^{*}).
\end{split}
\]
Denote $v^{*}=\left(\frac{1}{\lambda}\partial f_{1} \circ B\right)(u^{*})$. Then we obtain that

\begin{equation}\label{equ3.6}
u^{*}=u^{*}-Q^{-1}\nabla f_{2}(u^{*})-\lambda Q^{-1}B^{T}v^{*}.
\end{equation}
Besides, using Proposition \ref{pro1} we have
\begin{equation}\label{equ3.7}
\begin{split}
v^{*}=\left(\frac{1}{\lambda}\partial f_{1} \circ B\right)(u^{*})\Leftrightarrow
Bu^{*}=\textrm{prox}_{\frac{1}{\lambda}f_{1}}(Bu^{*}+v^{*}) \\
\Leftrightarrow v^{*}=\left(I-\textrm{prox}_{\frac{1}{\lambda}f_{1}}\right)(Bu^{*}+v^{*}).
\end{split}
\end{equation}
Inserting (\ref{equ3.6}) into (\ref{equ3.7}) we further obtain that
\begin{equation}\label{equ3.8}
v^{*}=\left(I-\textrm{prox}_{\frac{1}{\lambda}f_{1}}\right)(B(u^{*}-Q^{-1}\nabla f_{2}(u^{*}))+(I-\lambda BQ^{-1}B^{T})v^{*}).
\end{equation}
Based on (\ref{equ3.6}) and (\ref{equ3.8}) we infer that $(v^{*},u^{*})$ is a fixed point of $T$.

Conversely, if there exists $w^{*}=(v^{*},u^{*})$ satisfying $w^{*}=T(w^{*})$, we can derive that $u^{*}$ satisfies the first-order optimality condition of (\ref{equ1.1}) by the equivalent formulas above. Therefore, we conclude that $u^{*}$ is a solution of (\ref{equ1.1}).
\end{pf}

According to the formulas in (\ref{equ3.3}) and (\ref{equ3.4}) we find out that the iterative scheme of $\textrm{FP^{2}O\_QN}$ can be reformulated as
\[
\left\{\begin{array}{lll}v^{k+1} = T_{1}(v^{k}, u^{k}), &~ &~ ~\\
u^{k+1} = T_{2}(v^{k}, u^{k})
\end{array} \right.
\]
which is also equal to $w^{k+1}=T(w^{k})$ with $w^{k}=(v^{k}, u^{k})$. This implies that the sequence $\{v^{k}, u^{k}\}$ generated by $\textrm{FP^{2}O\_QN}$ is just the Picard sequence of the operator $T$. With the similar discussion we can find that the iterative formulas of $\textrm{FP^{2}O_{\kappa}\_QN}$ is equal to $w^{k+1}=T_{}\kappa(w^{k})$, i.e., the sequence $\{v^{k}, u^{k}\}$ generated by $\textrm{FP^{2}O_{\kappa}\_QN}$ is the Picard sequence of the operator $T_{\kappa}$.

Based on Theorem \ref{the1} we know that the solution of the minimization problem (\ref{equ1.1}) is just equal to the fixed point of the operator $T$. Therefore, the convergence of $\textrm{FP^{2}O_{\kappa}\_QN}$ can be guaranteed by verifying the nonexpansion of $T$ according to Proposition \ref{pro2}. The proof here is similar to those presented in \cite{IP:PDFP2O}. However, note that the global convergence of proposed algorithms cannot be directly obtained by applying results in \cite{IP:PDFP2O}, and hence it is included here for completion.

In the following, we give a crucial inequality for showing the nonexpansion of $T$. Denote
\[
h(u)=u-Q^{-1}\nabla f_{2}(u),
\]
\[
M=I-\lambda BQ^{-1}B^{T}.
\]
Here we assume that $0<\lambda\leq 1/\lambda_{\max}(BQ^{-1}B^{T})$, and hence $M$ is a symmetric positive semi-definite matrix. Therefore, we can define the semi-norm $\|u\|_{M}=\sqrt{u^{T}Mu}$, and then define the norm
\[
\|w\|_{\lambda, Q}=\sqrt{\|u\|^{2}_{Q}+\lambda \|v\|^{2}_{2}}.
\]

\begin{lemma}\label{lem1}
For any two points $w_{1}=(v_{1}, u_{1})$ and $w_{2}=(v_{2}, u_{2})$ in $\mathbb{R}^{M}\times \mathbb{R}^{N}$, the following inequality
\begin{multline}\label{equ3.9}
\|T(w_{1})-T(w_{2})\|^{2}_{\lambda, Q}\leq \|w_{1}-w_{2}\|^{2}_{\lambda, Q}-\|\nabla f_{2}(u_{1})-\nabla f_{2}(u_{2})\|^{2}_{2\beta-Q^{-1}} \\
-\|v_{1}-v_{2}\|^{2}_{I-M}-\lambda\|T_{1}(w_{1})-T_{1}(w_{2})-(v_{1}-v_{2})\|^{2}_{M}
\end{multline}
comes into existence.
\end{lemma}

\begin{pf}
According to Lemma~2.4 of \cite{SIAM:PFBS} we know that $I-prox_{\frac{1}{\lambda}f}$ is firmly nonexpansive, and hence obtain

\begin{equation}\label{equ3.10}
\begin{split}
\|T_{1}(w_{1})-T_{1}(w_{2})\|_{2}^{2}&\leq \langle T_{1}(w_{1})-T_{1}(w_{2}), B(h(u_{1})-h(u_{2}))+M(v_{1}-v_{2})\rangle \\
&=\langle T_{1}(w_{1})-T_{1}(w_{2}), B(h(u_{1})-h(u_{2}))\rangle + \langle T_{1}(w_{1})-T_{1}(w_{2}), M(v_{1}-v_{2})\rangle.
\end{split}
\end{equation}
Following the definition in (\ref{equ3.4}) we also have

\begin{equation}\label{equ3.11}
\begin{split}
&\|T_{2}(w_{1})-T_{2}(w_{2})\|_{Q}^{2} \\
&= \|h(u_{1})-h(u_{2})-\lambda Q^{-1}B^{T}(T_{1}(w_{1})-T_{1}(w_{2}))\|_{Q}^{2} \\
&= \|h(u_{1})-h(u_{2})\|_{Q}^{2}-2\lambda\langle h(u_{1})-h(u_{2}), B^{T}(T_{1}(w_{1})-T_{1}(w_{2}))\rangle \\
&+\lambda \|T_{1}(w_{1})-T_{1}(w_{2})\|_{2}^{2}-\lambda \|T_{1}(w_{1})-T_{1}(w_{2})\|_{M}^{2}.
\end{split}
\end{equation}
According to (\ref{equ3.10}) and (\ref{equ3.11}) we further have

\begin{equation}\label{equ3.12}
\begin{split}
&\|T(w_{1})-T(w_{2})\|^{2}_{\lambda, Q} \\
&= \|T_{2}(w_{1})-T_{2}(w_{2})\|_{Q}^{2} + \lambda\|T_{1}(w_{1})-T_{1}(w_{2})\|_{2}^{2}\\
&= \|h(u_{1})-h(u_{2})\|_{Q}^{2}-\lambda \|T_{1}(w_{1})-T_{1}(w_{2})\|_{M}^{2} \\
&+ 2\lambda\|T_{1}(w_{1})-T_{1}(w_{2})\|_{2}^{2}-2\lambda\langle h(u_{1})-h(u_{2}), B^{T}(T_{1}(w_{1})-T_{1}(w_{2}))\rangle\\
&\leq \|h(u_{1})-h(u_{2})\|_{Q}^{2}-\lambda \|T_{1}(w_{1})-T_{1}(w_{2})\|_{M}^{2} + 2\lambda \langle T_{1}(w_{1})-T_{1}(w_{2}), M(v_{1}-v_{2})\rangle \\
&= \|h(u_{1})-h(u_{2})\|_{Q}^{2}+\lambda \|v_{1}-v_{2}\|_{M}^{2}-\lambda \|T_{1}(w_{1})-T_{1}(w_{2})-(v_{1}-v_{2})\|_{M}^{2}.
\end{split}
\end{equation}
Since $f_{2}$ has $1/\beta$-Lipschitz continuous gradient, we get that

\begin{equation}\label{equ3.13}
\begin{split}
&\|h(u_{1})-h(u_{2})\|_{Q}^{2} \\
& = \|u_{1}-u_{2}\|_{Q}^{2}-2\langle \nabla f_{2}(u_{1})-\nabla f_{2}(u_{2}), u_{1}-u_{2}\rangle+\|\nabla f_{2}(u_{1})-\nabla f_{2}(u_{2})\|_{Q^{-1}}^{2} \\
&\leq \|u_{1}-u_{2}\|_{Q}^{2}-\|\nabla f_{2}(u_{1})-\nabla f_{2}(u_{2})\|_{2\beta-Q^{-1}}^{2}.
\end{split}
\end{equation}
By the definition of $M$ we easily get

\begin{equation}\label{equ3.14}
\|v_{1}-v_{2}\|_{M}^{2}=\|v_{1}-v_{2}\|^{2}_{2}-\|v_{1}-v_{2}\|^{2}_{I-M}.
\end{equation}
Based on (\ref{equ3.12})--(\ref{equ3.14}) we can obtain (\ref{equ3.9}) directly.
\end{pf}

From the results in Lemma \ref{lem1} we know that $T$ is nonexpansive with the norm of $\|\cdot\|_{\lambda, Q}$. Therefore, we are able to prove the convergence of $\textrm{FP^{2}O_{\kappa}\_QN}$ according to Proposition \ref{pro2}, which is described as follows.

\begin{theorem}\label{the2}
Assume that $\|Q^{-1}\|_{2}< 2\beta$ and $0< \lambda\leq 1/\lambda_{\max}(BQ^{-1}B^{T})$. Let $(v^{k}, u^{k})$ be the sequence generated by $\textrm{FP^{2}O_{\kappa}\_QN}$. Then $(v^{k}, u^{k})$ converges to the fixed point of $T$ and $u^{k}$ converges to the solution of (\ref{equ1.1}).
\end{theorem}

\begin{pf}
Note that the solution of (\ref{equ1.1}) is just one fixed point of $T$. From Lemma \ref{lem1} we know that the operator $T$ is nonexpansive, maps the set $\mathbb{R}^{M}\times \mathbb{R}^{N}$ to itself, and has at least one fixed point. According to Opial $\kappa$-averaged Theorem, we conclude that, for any $w^{0}$ and $\kappa\in (0,1)$, the Picard sequence of $T_{\kappa}$ converges to a fixed point of $T$. With this result we further infer that $u^{k}$ converges to the solution of (\ref{equ1.1}).
\end{pf}

Finally, we process with the convergence of $\textrm{FP^{2}O\_QN}$ based on the inequality in Lemma \ref{lem1}.

\begin{theorem}\label{the3}
Assume that $\|Q^{-1}\|_{2}< 2\beta$ and $0< \lambda\leq 1/\lambda_{\max}(BQ^{-1}B^{T})$. Let $(v^{k}, u^{k})$ be the sequence generated by $\textrm{FP^{2}O\_QN}$. Then $(v^{k}, u^{k})$ converges to the fixed point of $T$ and $u^{k}$ converges to the solution of (\ref{equ1.1}).
\end{theorem}

\begin{pf}
Let $w^{*}=(v^{*}, u^{*})\in \mathbb{R}^{M}\times \mathbb{R}^{N}$ be a fixed point of $T$. Substitute $w_{1}$ and $w_{2}$ in (\ref{equ3.9}) with $w^{k}=(v^{k}, u^{k})$ and $w^{*}$, we obtain that

\begin{equation}\label{equ3.15}
\|w^{k+1}-w^{*}\|^{2}_{\lambda, Q}\leq \|w^{k}-w^{*}\|^{2}_{\lambda, Q}-\|\nabla f_{2}(u^{k})-\nabla f_{2}(u^{*})\|^{2}_{2\beta-Q^{-1}}
-\|v^{k}-v^{*}\|^{2}_{I-M}-\lambda\|v^{k+1}-v^{k}\|^{2}_{M}.
\end{equation}
Summing (\ref{equ3.15}) from some $k_{0}$ to $+\infty$ we obtain that

\begin{equation}\label{equ3.16}
\sum_{k=k_{0}}^{+\infty}\left\{\|\nabla f_{2}(u^{k})-\nabla f_{2}(u^{*})\|^{2}_{2\beta-Q^{-1}}
+\|v^{k}-v^{*}\|^{2}_{I-M}+\lambda\|v^{k+1}-v^{k}\|^{2}_{M}\right\}\leq \|w^{k_{0}}-w^{*}\|^{2}_{\lambda, Q}
\end{equation}
which implies that

\begin{equation}\label{equ3.17}
\lim\limits_{k\rightarrow +\infty}\|\nabla f_{2}(u^{k})-\nabla f_{2}(u^{*})\|^{2}_{2\beta-Q^{-1}}=0,
\end{equation}

\begin{equation}\label{equ3.18}
\lim\limits_{k\rightarrow +\infty}\|v^{k}-v^{*}\|^{2}_{I-M}=0,
\end{equation}

\begin{equation}\label{equ3.19}
\lim\limits_{k\rightarrow +\infty}\|v^{k+1}-v^{k}\|^{2}_{M}=0.
\end{equation}

According to (\ref{equ3.18}) we can easily deduce that $\lim\limits_{k\rightarrow +\infty}\|v^{k+1}-v^{k}\|^{2}_{I-M}=0$, and combining it with (\ref{equ3.19}) we have

\begin{equation}\label{equ3.20}
\lim\limits_{k\rightarrow +\infty}\|v^{k+1}-v^{k}\|^{2}_{2}=0.
\end{equation}
Besides, from (\ref{equ3.6}) we know that $Q^{-1}\nabla f_{2}(u^{*})+\lambda Q^{-1}B^{T}v^{*}=0$, and hence obtain that

\begin{equation}\label{equ3.21}
u^{k+1}-u^{k}=-Q^{-1}(\nabla f_{2}(u^{k})-\nabla f_{2}(u^{*}))-\lambda Q^{-1}B^{T}(v^{k+1}-v^{*}).
\end{equation}
Based on (\ref{equ3.21}) we immediately get

\begin{equation}\label{equ3.22}
\begin{split}
\|u^{k+1}-u^{k}\|^{2}_{Q}
&\leq \|Q^{-1}(\nabla f_{2}(u^{k})-\nabla f_{2}(u^{*}))\|^{2}_{Q}+\|\lambda Q^{-1}B^{T}(v^{k+1}-v^{*})\|^{2}_{Q}\\
&= \|(\nabla f_{2}(u^{k})-\nabla f_{2}(u^{*}))\|^{2}_{Q^{-1}}+\lambda \|v^{k+1}-v^{*}\|^{2}_{I-M}.
\end{split}
\end{equation}
Since $0<\|Q^{-1}\|_{2}< 2\beta$, utilizing (\ref{equ3.17}) and (\ref{equ3.18}) we can deduce from (\ref{equ3.22}) that

\begin{equation}\label{equ3.23}
\lim\limits_{k\rightarrow +\infty}\|u^{k+1}-u^{k}\|^{2}_{Q}=0.
\end{equation}
Combining (\ref{equ3.20}) and (\ref{equ3.23}) we further obtain that

\begin{equation}\label{equ3.24}
\lim\limits_{k\rightarrow +\infty}\|w^{k+1}-w^{k}\|_{\lambda, Q}=0.
\end{equation}
According to (\ref{equ3.15}) we know that the sequence $\left\{\|w^{k}-w^{*}\|^{2}_{\lambda, Q}\right\}$ is non-increasing, and hence $\left\{w^{k}\right\}$ is bounded, which implies that there exists a convergent subsequence of $\left\{w^{k_{j}}\right\}$ such that

\begin{equation}\label{equ3.25}
\lim\limits_{j\rightarrow +\infty}\|w^{k_{j}}-\hat{w}^{*}\|_{\lambda, Q}=0
\end{equation}
for some point $\hat{w}^{*}=(\hat{v}^{*}, \hat{u}^{*})$. Due to the operator $T$ is continuous, we further have $\lim\limits_{j\rightarrow +\infty}\|T(w^{k_{j}})-T(\hat{w}^{*})\|_{\lambda, Q}=0$. Besides, we have

\begin{equation}\label{equ3.26}
\|T(w^{k_{j}})-\hat{w}^{*}\|_{\lambda, Q}\leq \|w^{k_{j}+1}-w^{k_{j}}\|_{\lambda, Q}+\|w^{k_{j}}-\hat{w}^{*}\|_{\lambda, Q}
\end{equation}
which implies that

\begin{equation}\label{equ3.27}
\lim\limits_{j\rightarrow +\infty}\|T(w^{k_{j}})-\hat{w}^{*}\|_{\lambda, Q}=0
\end{equation}
based on (\ref{equ3.24}) and (\ref{equ3.25}). Therefore, $\hat{w}^{*}$ is a fixed point of the operator $T$. Due to the proof is started with any fixed point $w^{*}$, we can set $w^{*}=\hat{w}^{*}$. In this case, we see that the sequence $\left\{\|w^{k}-\hat{w}^{*}\|_{\lambda, Q}\right\}$ is non-increasing. Combining this with the formula (\ref{equ3.25}) we infer that

\begin{equation}\label{equ3.28}
\lim\limits_{k\rightarrow +\infty}w^{k}=\hat{w}^{*}.
\end{equation}
From Theorem \ref{the1} we know that $\hat{u}^{*}$ is a solution of (\ref{equ1.1}). This completes the proof.
\end{pf}

\section{Numerical examples}\label{sec4}
\setcounter{equation}{0}

In this section, we will compare the proposed $\textrm{FP^{2}O_{\kappa}\_QN}$ algorithm with $\textrm{PDFP^{2}O_{\kappa}}$ \cite{IP:PDFP2O} through the experiments of image restoration. Here two cases of additive and multiplicative noise types are considered. One is the additive Gaussian noise which has been extensively investigated over the last decades. In this setting, the data fidelity term can be formulated as
\[
f_{2}(u)=\frac{1}{2}\|K u - b\|_{2}^{2}
\]
where $K$ is the blurring operator, and $b$ is the observed image. The other is the speckle noise which also appears in many real world image processing applications such as laser imaging, synthetic aperture radar (SAR) imaging and ultrasonic imaging. In \cite{CVPR:Rayleigh}, this speckle noise followed by a Rayleigh distribution is investigated. Under this condition, the observed image can be modeled as corrupted with signal-dependent noise of this form

\begin{equation}\label{equ4.1}
b = K u + \sqrt{K u}\varepsilon
\end{equation}
where $\varepsilon$ is a zero-mean Gaussian noise with standard deviation $\sigma$, i.e., $\varepsilon\sim N(0, \sigma)$. Based on the model (\ref{equ4.1}) and the characteristics of Gaussian distribution, the corresponding fidelity term can be formulated as
\[
f_{2}(u)=\sum_{i}\frac{(b-K u)_{i}^{2}}{(Ku)_{i}}.
\]

In the following experiments, we use total variation as the regularization term, and hence choose the function
\[
f_{1}(B u)=\mu\|\nabla u\|_{1}
\]
where $\nabla: \mathbb{R}^{N}\rightarrow \mathbb{R}^{2N}$ is a discrete gradient operator. Here we adopt the isotropic definition of total variation, and the proximity operator $\textrm{prox}_{\frac{1}{\lambda}f_{1}}$ can be computed easily. For more details refer to \cite{JMIV:IFP2O}.

\subsection{Gaussian image deblurring}\label{subsec4.1}

In this subsection, we choose three gray-scale images, Cameraman, Barbara (with size of $256\times256$), and Boat (with size of $512\times512$) as the original images, and evaluate $\textrm{FP^{2}O_{\kappa}\_QN}$ in four typical image blurring scenarios: strong blur with low noise; strong blur with medium noise; mild blur with low noise; mild blur with medium noise, which are summarized in Table \ref{tab4.1} ($\sigma$ and $\sigma_{a}$ denote the standard deviation).

\begin{table} [htbp]
\centering \caption{ Description of image blurring scenarios}
\scalebox{0.9}{
\begin{tabular}{c|c|c}
  \hline
  Scenario & Blur kernel & Gaussian noise \\
  \hline
  1 & $8\times 8$ box average kernel & $\sigma = 1.5$\\

  2 & $8\times 8$ box average kernel & $\sigma = 3$\\

  3 & $6\times 6$ gaussian kernel with $\sigma_{a}=8$ & $\sigma = 1.5$\\

  4 & $6\times 6$ gaussian kernel with $\sigma_{a}=8$ & $\sigma = 3$\\
  \hline
\end{tabular}}
\label{tab4.1}
\end{table}

In the following, we discuss the selection of the parameters $\mu$, $\lambda$ and $\kappa$ in both fixed point algorithms, the parameter $\gamma$ in $\textrm{PDFP^{2}O_{\kappa}}$, and the matrix $Q$ in $\textrm{FP^{2}O_{\kappa}\_QN}$. Due to $f_{2}(u)=\frac{1}{2}\|K u - b\|_{2}^{2}$, we have $\nabla^{2}f_{2}\equiv K^{T}K$. Therefore, we can choose $Q=K^{T}K$. However, the blurring operator $K$ is ill-posed generally, and $(K^{T}K)^{-1}$ cannot be used in the proposed algorithm due to the instability. Therefore, we choose $Q=K^{T}K +\epsilon B^{T}B$ in our experiments. Here $\epsilon$ is a small positive number and $B$ is a difference matrix. Notice that the introduction of the term $\epsilon B^{T}B$ avoids the ill-posed condition, and $Q^{-1}$ can also be computed efficiently by fast Fourier transforms (FFTs) with periodic boundary conditions.

The regularization parameter $\mu$ is decided by the noise level, and the adjustment of the parameters $\lambda$ and $\gamma$ does influence the convergence speed and stability of the fixed point algorithms. Through many trials we use the rules of thumb: $\mu$ is set to $0.06$ and $0.15$ for $\sigma=1.5$ and $3.0$ respectively; $\lambda$ is set to $0.125$; $\gamma$ is chosen to be $1.8$ for $\textrm{PDFP^{2}O_{\kappa}}$; and $\epsilon=0.1$ for $\textrm{FP^{2}O_{\kappa}\_QN}$. Similarly to the literatures \cite{JMIV:IFP2O, IP:PDFP2O}, we find that $\kappa=0$ achieves the best convergence speed compared with other $\kappa\in (0,1)$, and hence we choose $\kappa=0$ for both algorithms.

The performance of the restored images of the compared algorithms is measured quantitatively by means of the peak signal-to-noise ratio (PSNR), which is defined by

\begin{equation}\label{equ4.2}
\textrm{PSNR}(u,\bar{u})=10\lg
\left\{\frac{255^{2}N}{\|u-u^{*}\|_2^{2}}\right\}
\end{equation}
where $u$ and $u^{*}$ denote the original image and the restored image respectively. The stopping criterion for the fixed-point algorithms is defined such that the relative error is below some small constant, i.e.,

\begin{equation}\label{equ4.3}
\frac{\|u^{k+1}-u^{k}\|_{2}}{\|u^{k}\|_{2}}<tol
\end{equation}
where tol denotes a prescribed tolerance value. In our experiments we choose $tol=5\times 10^{-4}$.

\begin{figure}
  \centering
  \subfigure[]{
    \label{fig4.1:subfig:a} 
    \includegraphics[width=2.0in,clip]{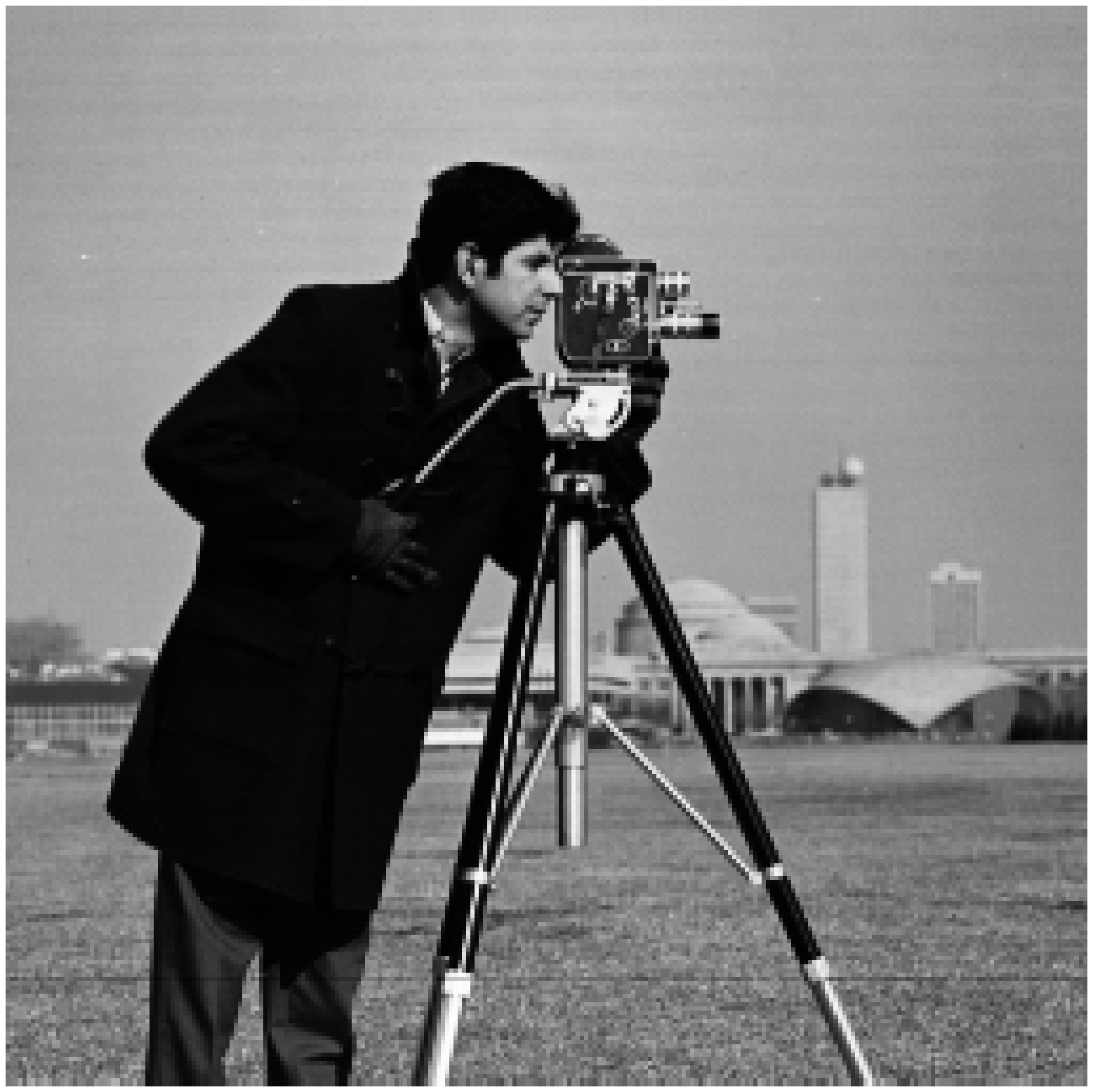}}
  \subfigure[]{
    \label{fig4.1:subfig:b} 
    \includegraphics[width=2.0in,clip]{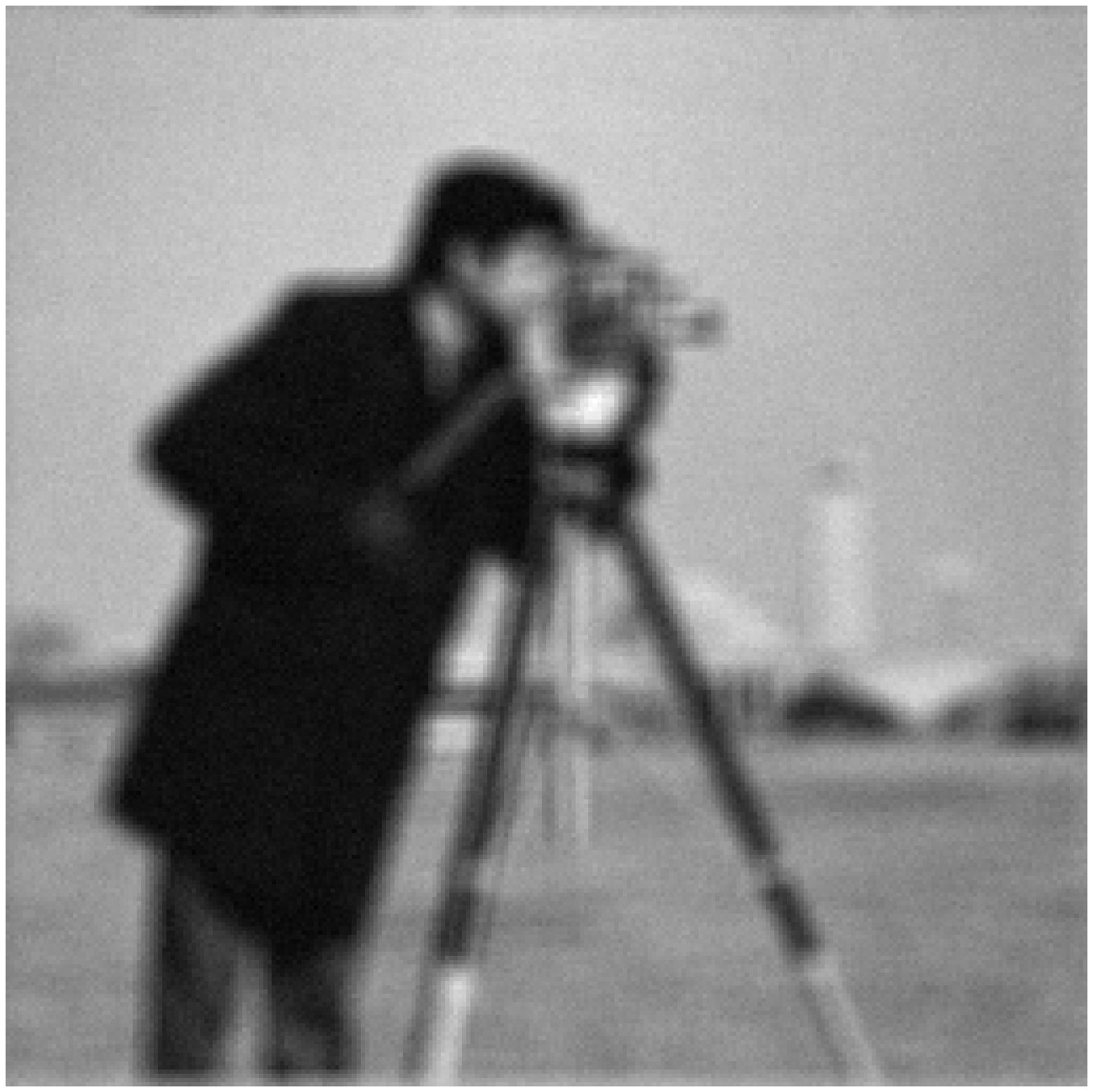}}
  \subfigure[]{
    \label{fig4.1:subfig:c} 
    \includegraphics[width=2.0in,clip]{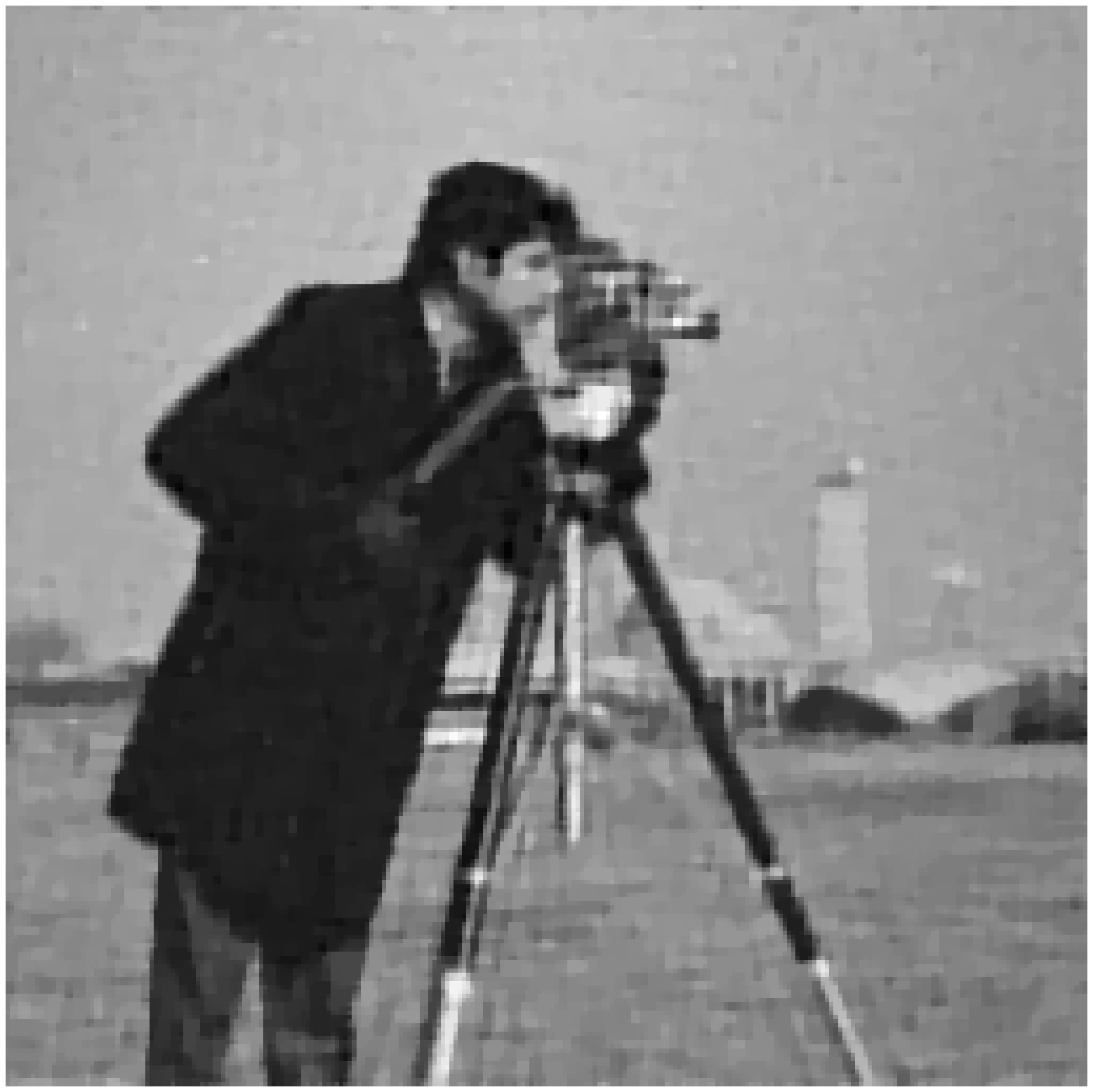}}
  \subfigure[]{
    \label{fig4.1:subfig:d} 
    \includegraphics[width=2.0in,clip]{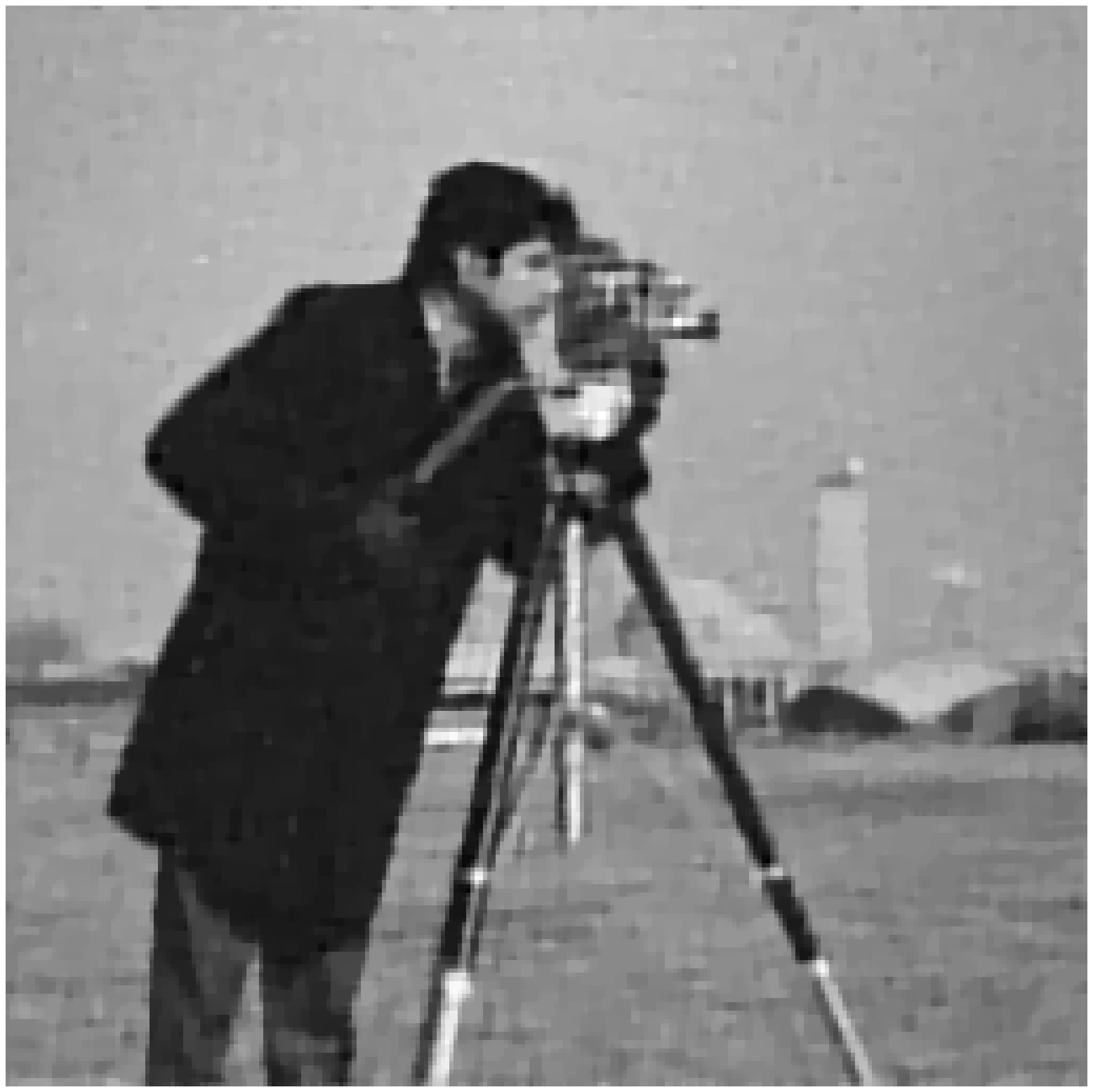}}
\caption{(a) The original Cameraman image, (b) the blurry and noisy image in the scenario 2, PSNR=20.96dB, (c) the image restored by $\textrm{PDFP^{2}O_{\kappa}}$, (d) the image restored by $\textrm{FP^{2}O_{\kappa}\_QN}$.
}
\label{fig4.1}
\end{figure}

\begin{figure}
  \centering
  \subfigure[]{
    \label{fig4.2:subfig:a} 
    \includegraphics[width=2.0in,clip]{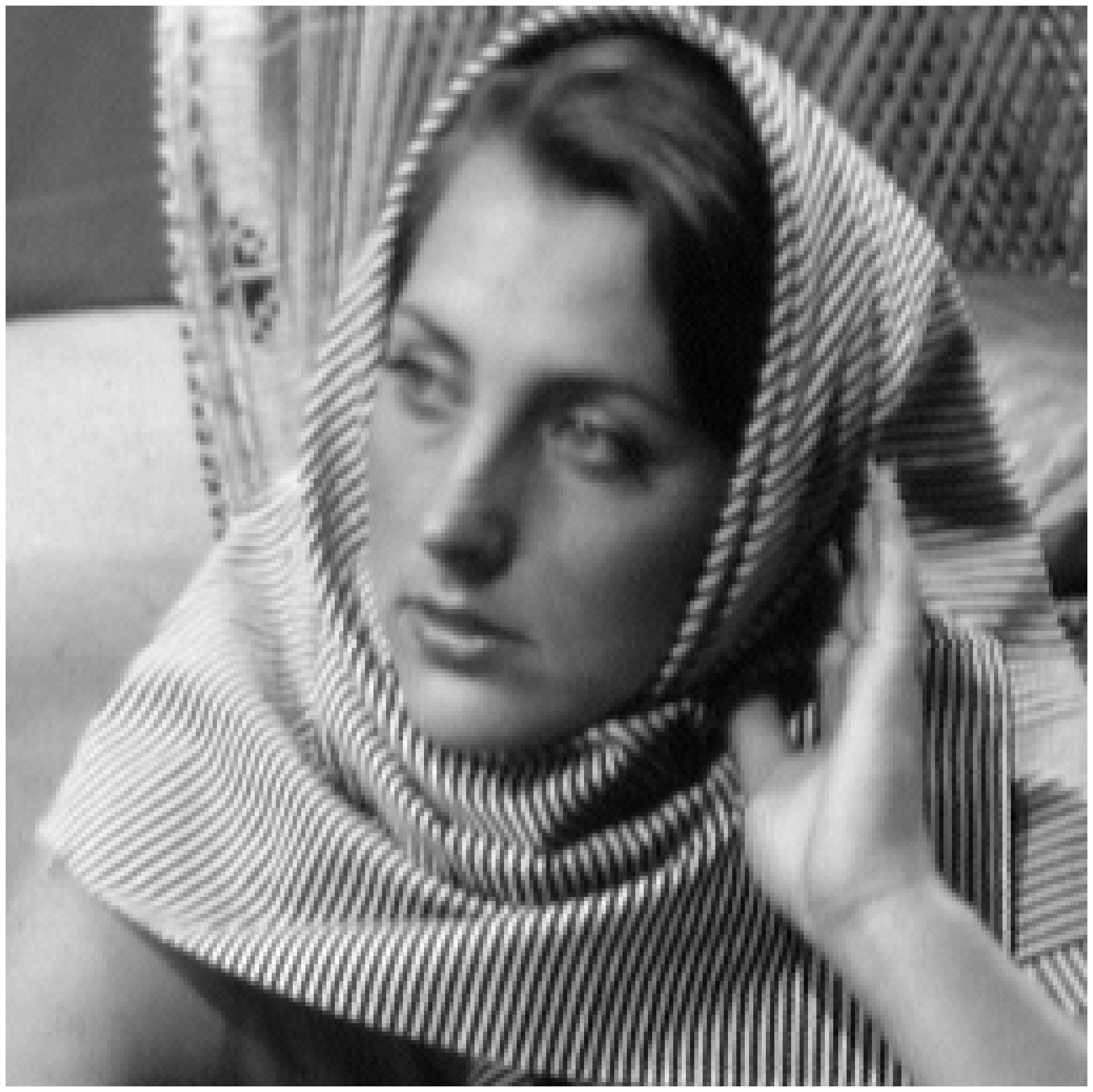}}
  \subfigure[]{
    \label{fig4.2:subfig:b} 
    \includegraphics[width=2.0in,clip]{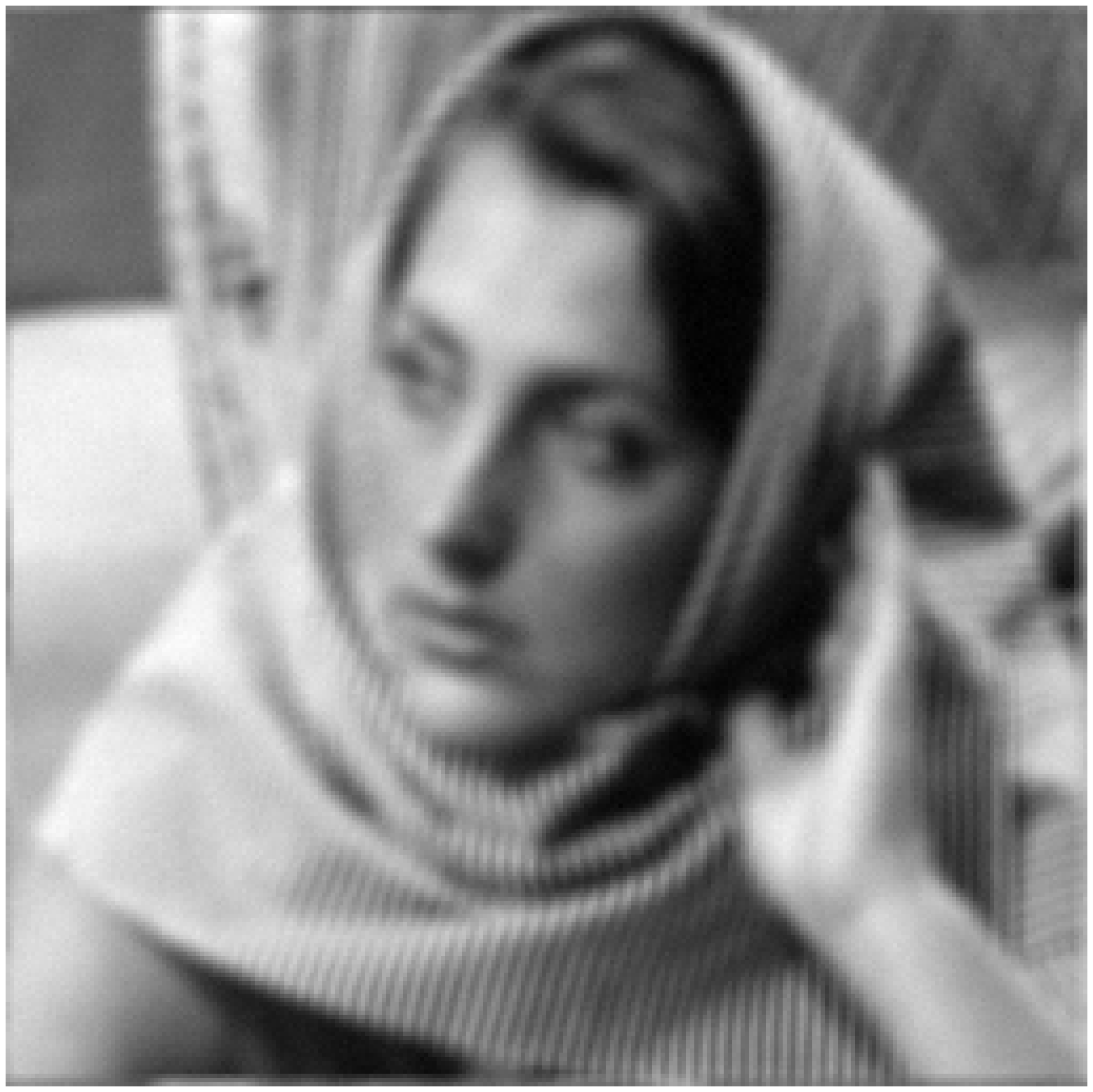}}
  \subfigure[]{
    \label{fig4.2:subfig:c} 
    \includegraphics[width=2.0in,clip]{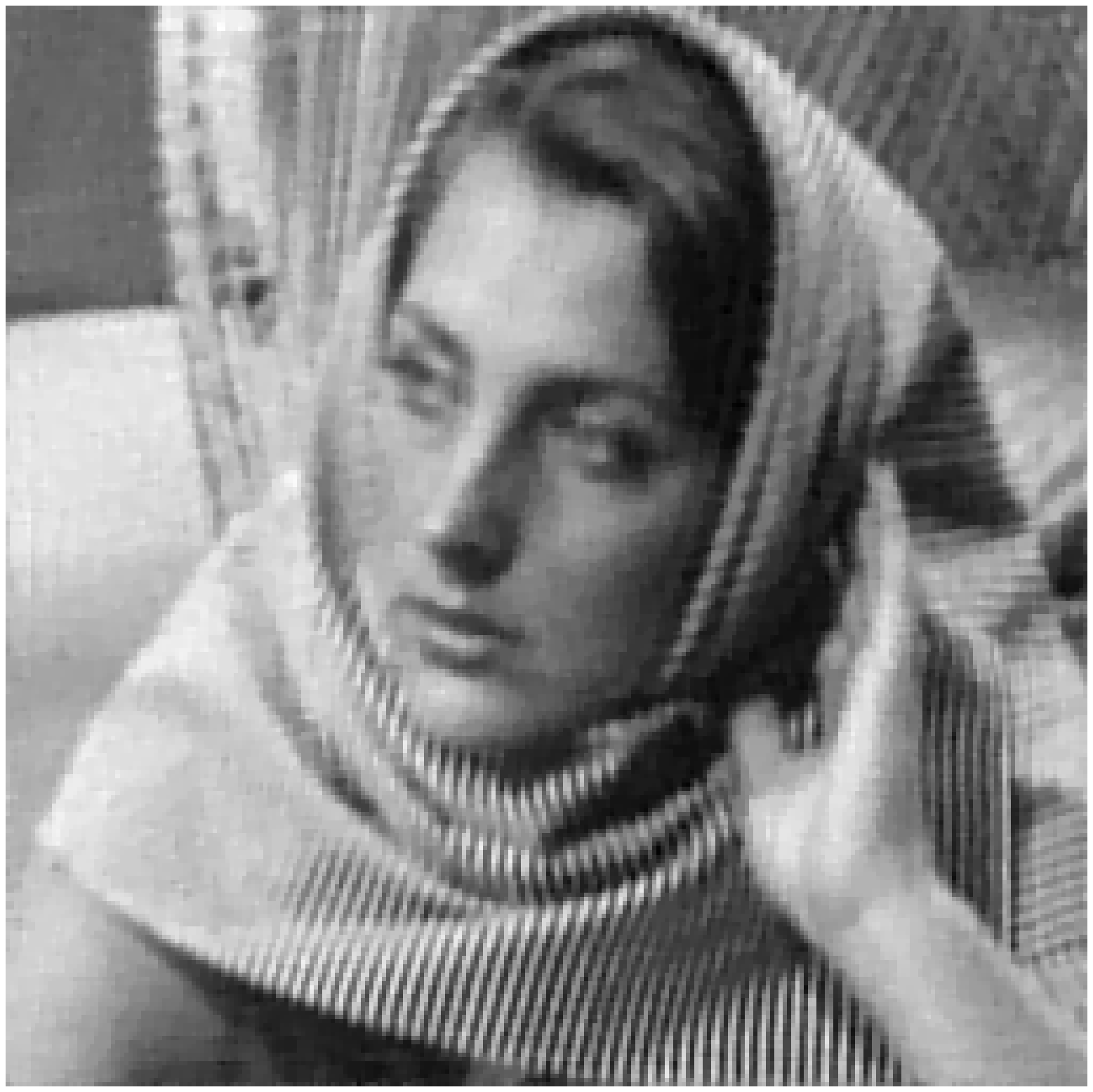}}
  \subfigure[]{
    \label{fig4.2:subfig:d} 
    \includegraphics[width=2.0in,clip]{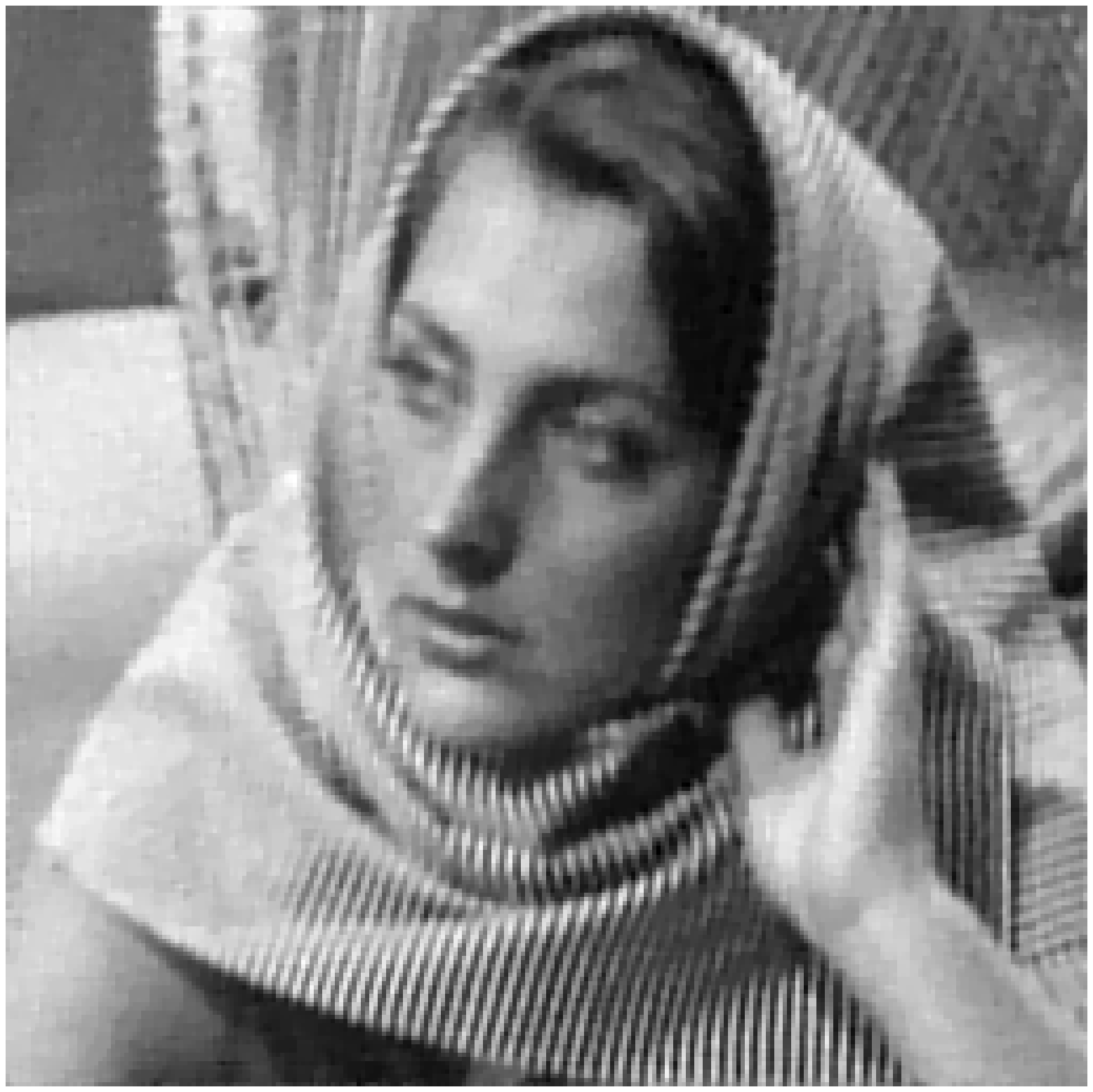}}
\caption{(a) The original Barbara image, (b) the blurry and noisy image in the scenario 3, PSNR=22.62dB, (c) the image restored by $\textrm{PDFP^{2}O_{\kappa}}$, (d) the image restored by $\textrm{FP^{2}O_{\kappa}\_QN}$.
}
\label{fig4.2}
\end{figure}

\begin{figure}
  \centering
  \subfigure[]{
    \label{fig4.3:subfig:a} 
    \includegraphics[width=2.0in,clip]{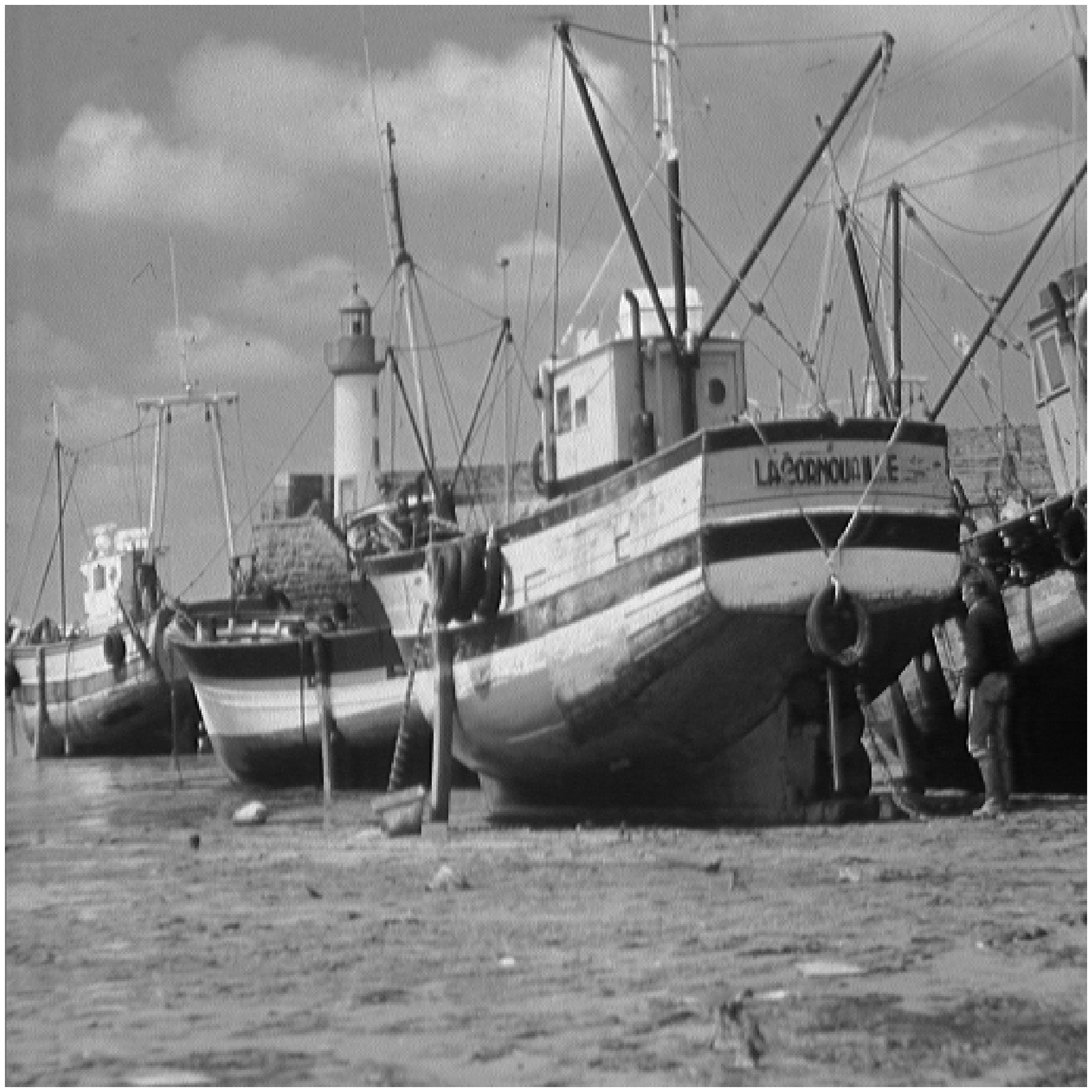}}
  \subfigure[]{
    \label{fig4.3:subfig:b} 
    \includegraphics[width=2.0in,clip]{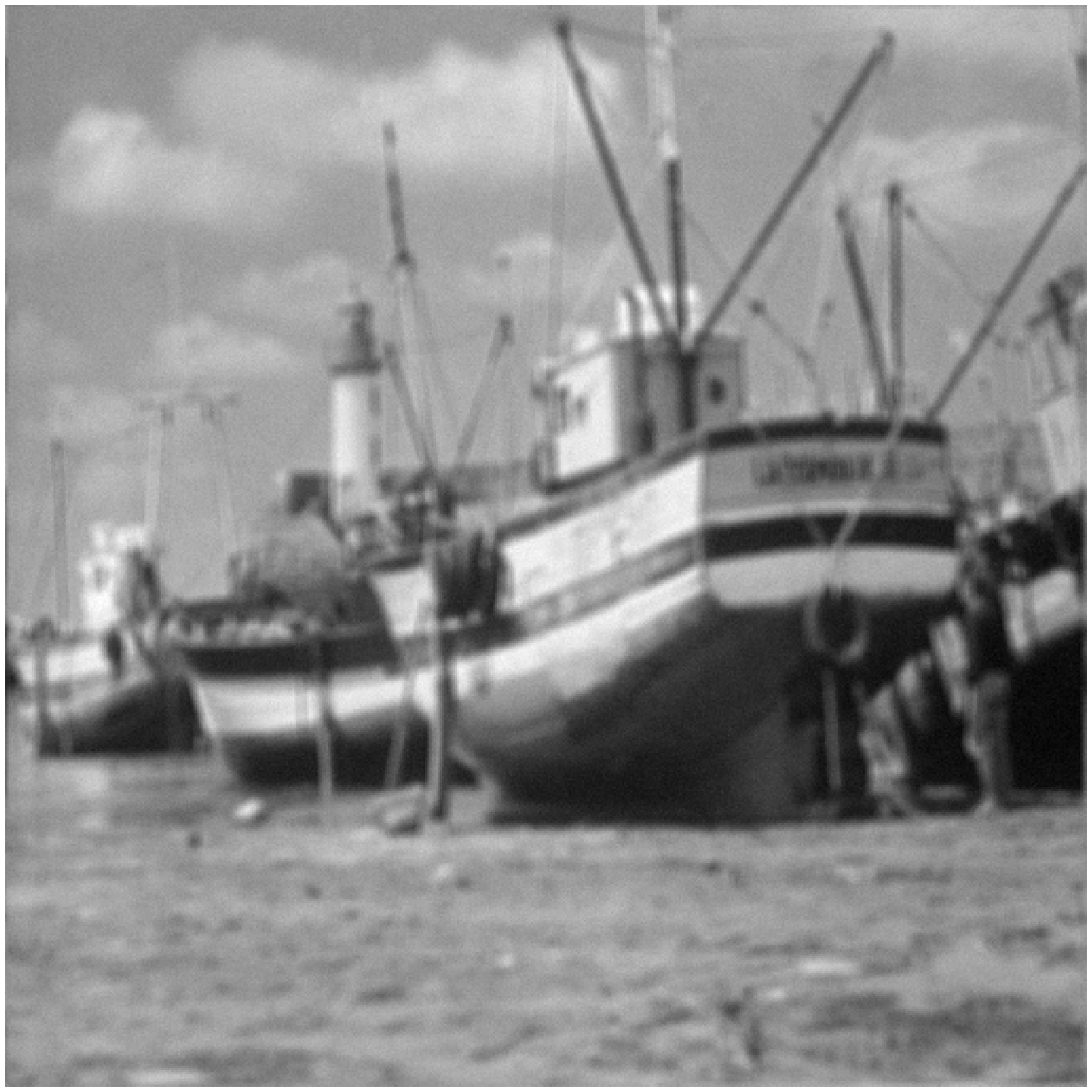}}
  \subfigure[]{
    \label{fig4.3:subfig:c} 
    \includegraphics[width=2.0in,clip]{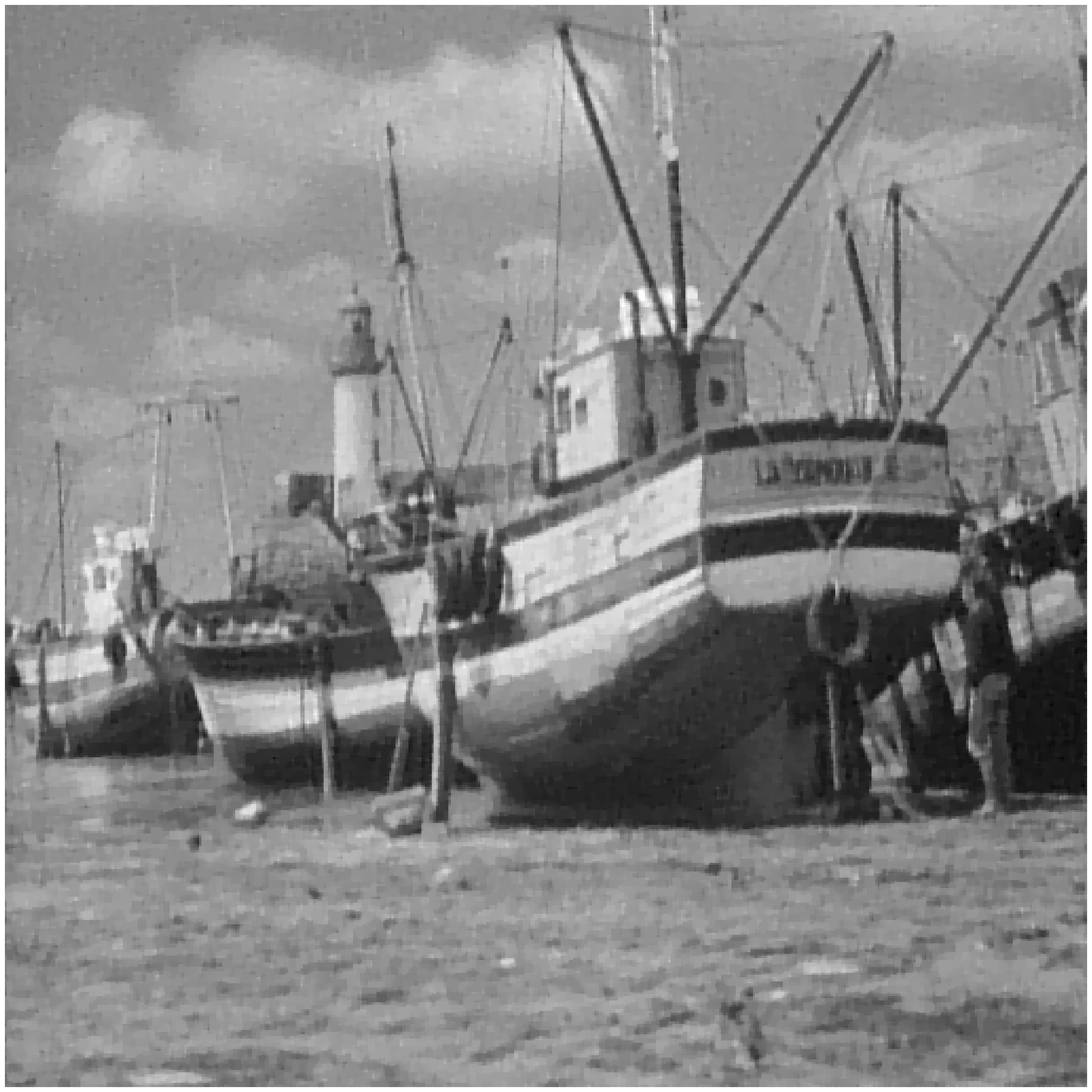}}
  \subfigure[]{
    \label{fig4.3:subfig:d} 
    \includegraphics[width=2.0in,clip]{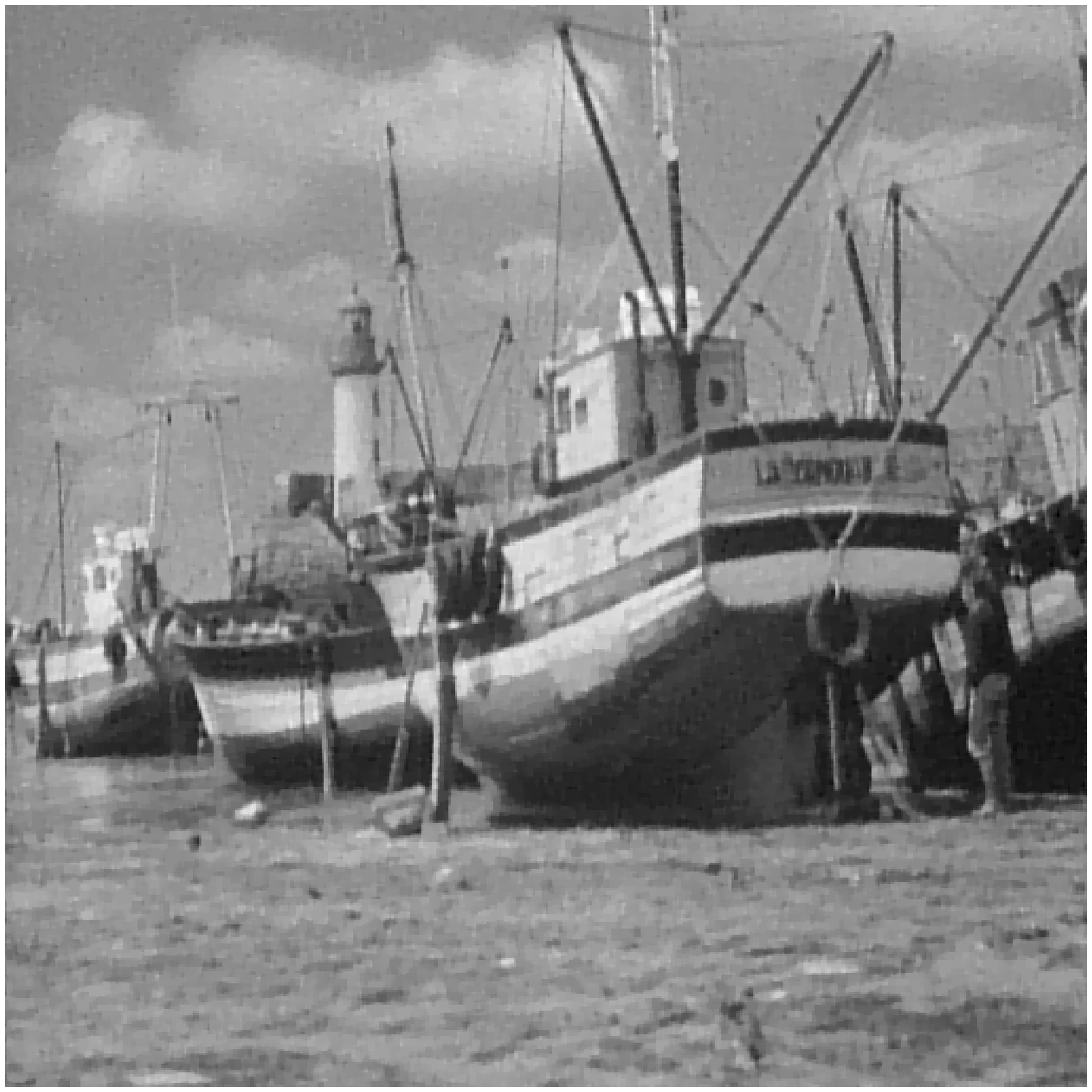}}
\caption{(a) The original Boat image, (b) the blurry and noisy image in the scenario 4, PSNR=24.74dB, (c) the image restored by $\textrm{PDFP^{2}O_{\kappa}}$, (d) the image restored by $\textrm{FP^{2}O_{\kappa}\_QN}$.
}
\label{fig4.3}
\end{figure}

The PSNR values for the deblurred images, the number of iterations, and the CPU time are listed in Table \ref{tab4.2}. In this table, the four image blurring scenarios shown in Table \ref{tab4.1} are considered, and $(\cdot, \cdot, \cdot )$ represents the PSNR values, iteration numbers and CPU time in sequence. From these results we observe that the recovered images obtained by $\textrm{FP^{2}O_{\kappa}\_QN}$ can achieve better PSNRs than those given by $\textrm{PDFP^{2}O_{\kappa}}$, and meanwhile, the corresponding iteration number and running time of $\textrm{FP^{2}O_{\kappa}\_QN}$ is less than those of $\textrm{PDFP^{2}O_{\kappa}}$. Figures \ref{fig4.1}--\ref{fig4.3} show the recovery results of the $\textrm{PDFP^{2}O_{\kappa}}$ and $\textrm{FP^{2}O_{\kappa}\_QN}$ algorithms. It is observed that the visual qualities of images obtained by both algorithms are more or less the same.

\begin{table} [htbp]
\centering \caption{The comparison of the performance of both fixed point algorithms: the given numbers are PSNR (dB)/Iteration number/CPU time(second) }
\scalebox{0.9}{
\begin{tabular}{ccccc}
  \hline
  Scenario & 1 & 2 & 3 & 4 \\
  \hline
  Image & \multicolumn{4}{c}{Cameraman} \\
  \hline
  $\textrm{PDFP^{2}O_{\kappa}}$ & (26.16, 97, 1.87) & (25.63, 102, 2.04) & (27.58, 89, 1.67) & (26.74, 88, 1.59)\\
  \hline
  $\textrm{FP^{2}O_{\kappa}\_QN}$ & (26.75, 46, 1.06) & (25.91, 42, 0.81) & (28.02, 45, 0.97) & (26.98, 42, 0.84)\\
  \hline
  Image & \multicolumn{4}{c}{Barbara} \\
  \hline
  $\textrm{PDFP^{2}O_{\kappa}}$ & (25.11, 73, 1.42) & (24.35, 74, 1.49) & (27.90, 75, 1.44) & (26.60, 75, 1.45)\\
  \hline
  $\textrm{FP^{2}O_{\kappa}\_QN}$ & (25.29, 38, 0.78) & (24.40, 34, 0.72) & (28.06, 39, 0.75) & (26.65, 38, 0.74)\\
   \hline
  Image & \multicolumn{4}{c}{Boat} \\
  \hline
  $\textrm{PDFP^{2}O_{\kappa}}$ & (28.64, 73, 5.27) & (27.83, 78, 6.09) & (30.11, 65, 4.98) & (28.95, 69, 5.19)\\
  \hline
  $\textrm{FP^{2}O_{\kappa}\_QN}$ & (29.35, 34, 3.57) & (28.15, 32, 3.31) & (30.64, 33, 3.28) & (29.20, 32, 3.24)\\
   \hline
\end{tabular}}
\label{tab4.2}
\end{table}

\begin{figure}
  \centering
  \subfigure[]{
    \label{fig4.4:subfig:a} 
    \includegraphics[width=2.5in,clip]{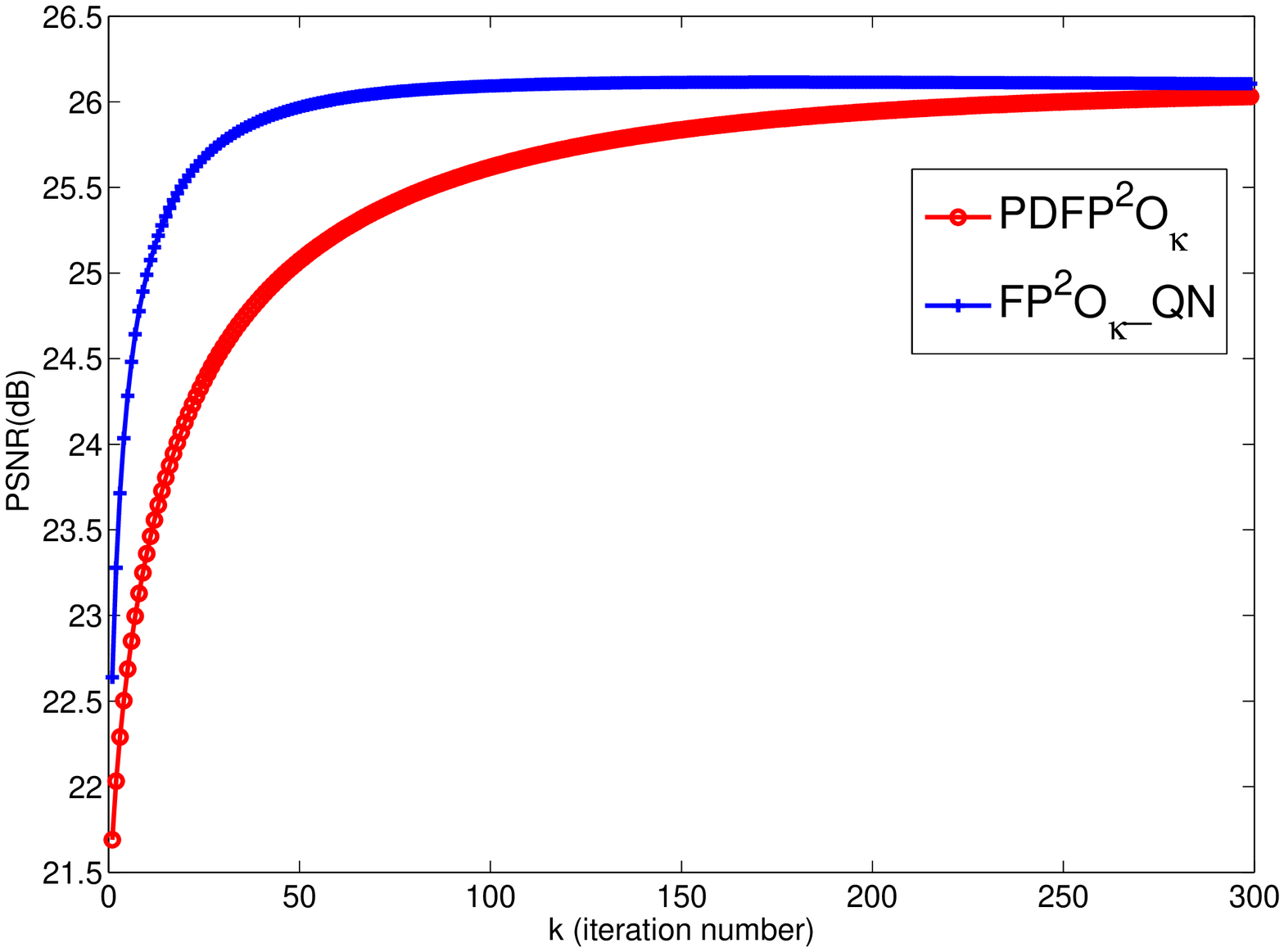}}
  \subfigure[]{
    \label{fig4.4:subfig:b} 
    \includegraphics[width=2.45in,clip]{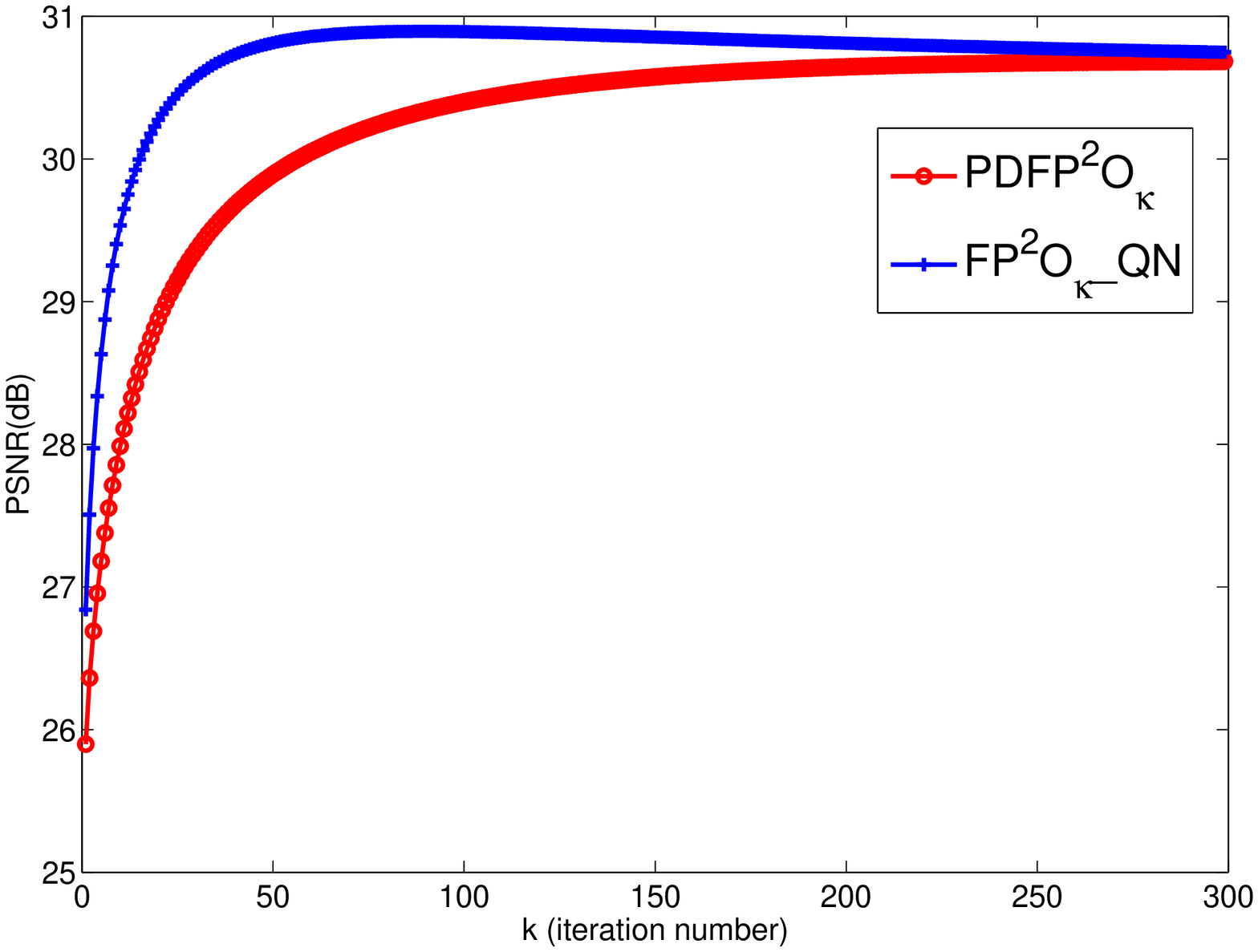}}
\caption{The evolution curves of PSNR (dB) for images with different blur kernels and noise levels. (a) Cameraman image in the scenario 2; (b) Boat image in the scenario 3.
}
\label{fig4.4}
\end{figure}

Figure \ref{fig4.4} shows the evolution curves of PSNR values obtained by both fixed point algorithms for two cases including in Table \ref{tab4.2}: one is the Cameraman image blurred by $8\times 8$ box average kernel and added with Gaussain noise with $\sigma = 3$, the other is the Boat image blurred by $6\times 6$ gaussian kernel and added with Gaussain noise with $\sigma = 1.5$. From the plots we can implicitly find that $\textrm{FP^{2}O_{\kappa}\_QN}$ achieves the best solution (with higher PSNRs) much faster than $\textrm{PDFP^{2}O_{\kappa}}$.

\subsection{Rayleigh image deblurring}\label{subsec4.2}

In this subsection, we further discuss the case of images contaminated by Rayleigh noise. The corresponding minimization problem has been introduced above. The two blur kernels shown in Table \ref{tab4.1}, and Rayleigh noise with $\sigma=0.5$ and $1.0$ are considered here.

First of all, we illustrate the setting of the parameters in both fixed point algorithms. Since $f_{2}(u)=\sum_{i}\frac{(b-K u)_{i}^{2}}{(Ku)_{i}}$, we have that
\[
\nabla^{2}f_{2}(\xi)=2K^{T}\left(\frac{b^{2}}{(K\xi)^{3}}\right)K
\]
for any $\xi\in \mathbb{R}^{N}$. Notice that the value of $\nabla^{2}f_{2}$ changes with the iteration number, and the inverse of $\nabla^{2}f_{2}$ is difficult to be estimated. Therefore, we use $Q=\beta K^{T}K+\epsilon$ to approximate $\nabla^{2}f_{2}$ in the proposed fixed point algorithm. Here the parameter $\beta$ is used to replace the unknown $\frac{b^{2}}{(K\xi)^{3}}$, and the term $\epsilon B^{T}B$ is included to avoid the ill-posed condition. In the following experiments, we find that $\beta=0.25$ and $\epsilon=0.005$ are two suitable selection through many trials. Moreover, for parameters in both algorithms we use the following rules of thumb: the regularization parameter $\mu$ is chosen to be $0.01$ and $0.02$ for the noise level of $\sigma=0.5$ and $1.0$ respectively; $\lambda$ is set to $0.125$; $\gamma$ is chosen to be $15.0$ for $\textrm{PDFP^{2}O_{\kappa}}$. We also find out that the selection of $\kappa=0$ is suitable for our experiments here.

\begin{figure}
  \centering
  \subfigure[]{
    \label{fig4.5:subfig:a} 
    \includegraphics[width=2.5in,clip]{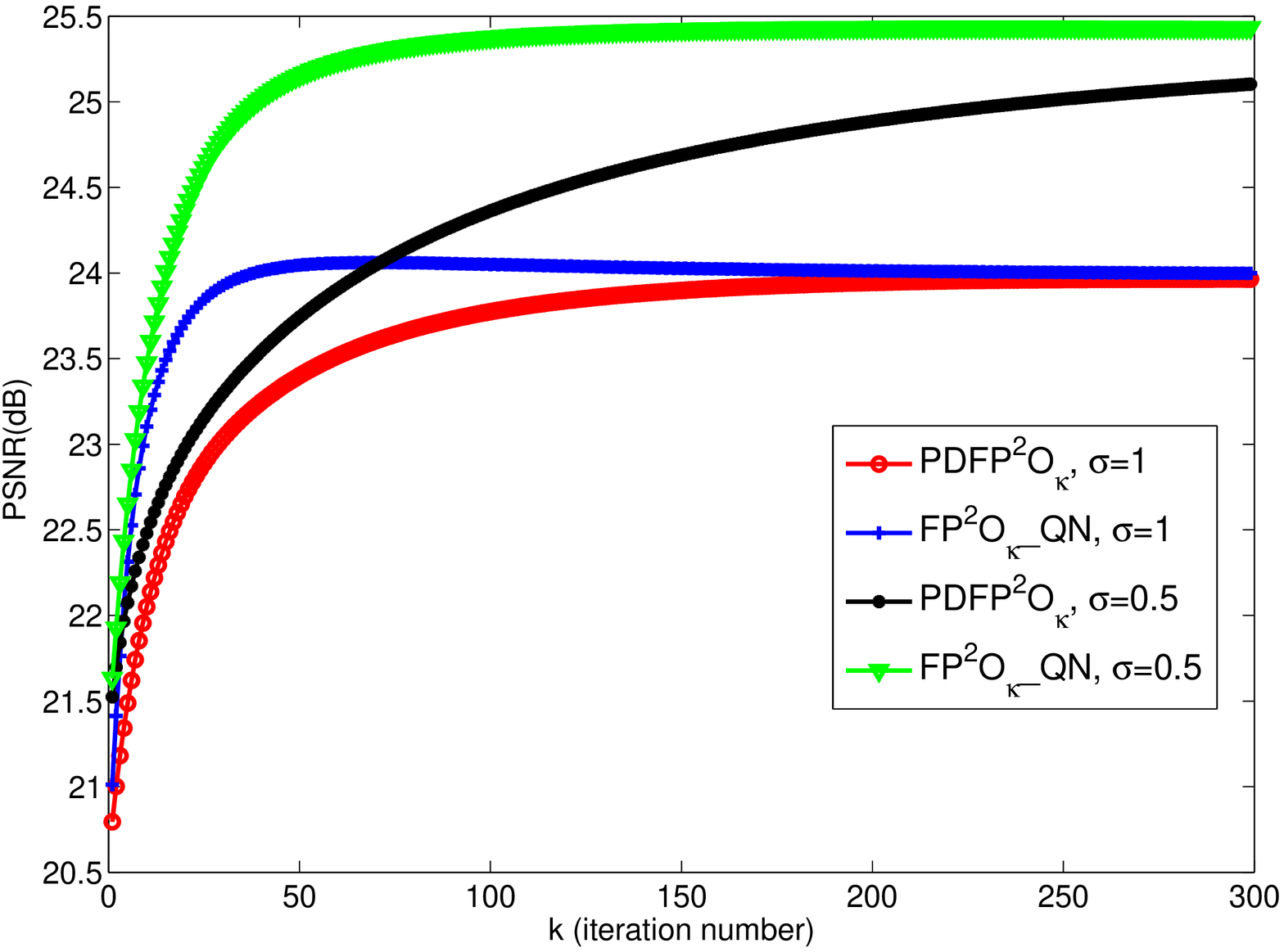}}
  \subfigure[]{
    \label{fig4.5:subfig:b} 
    \includegraphics[width=2.45in,clip]{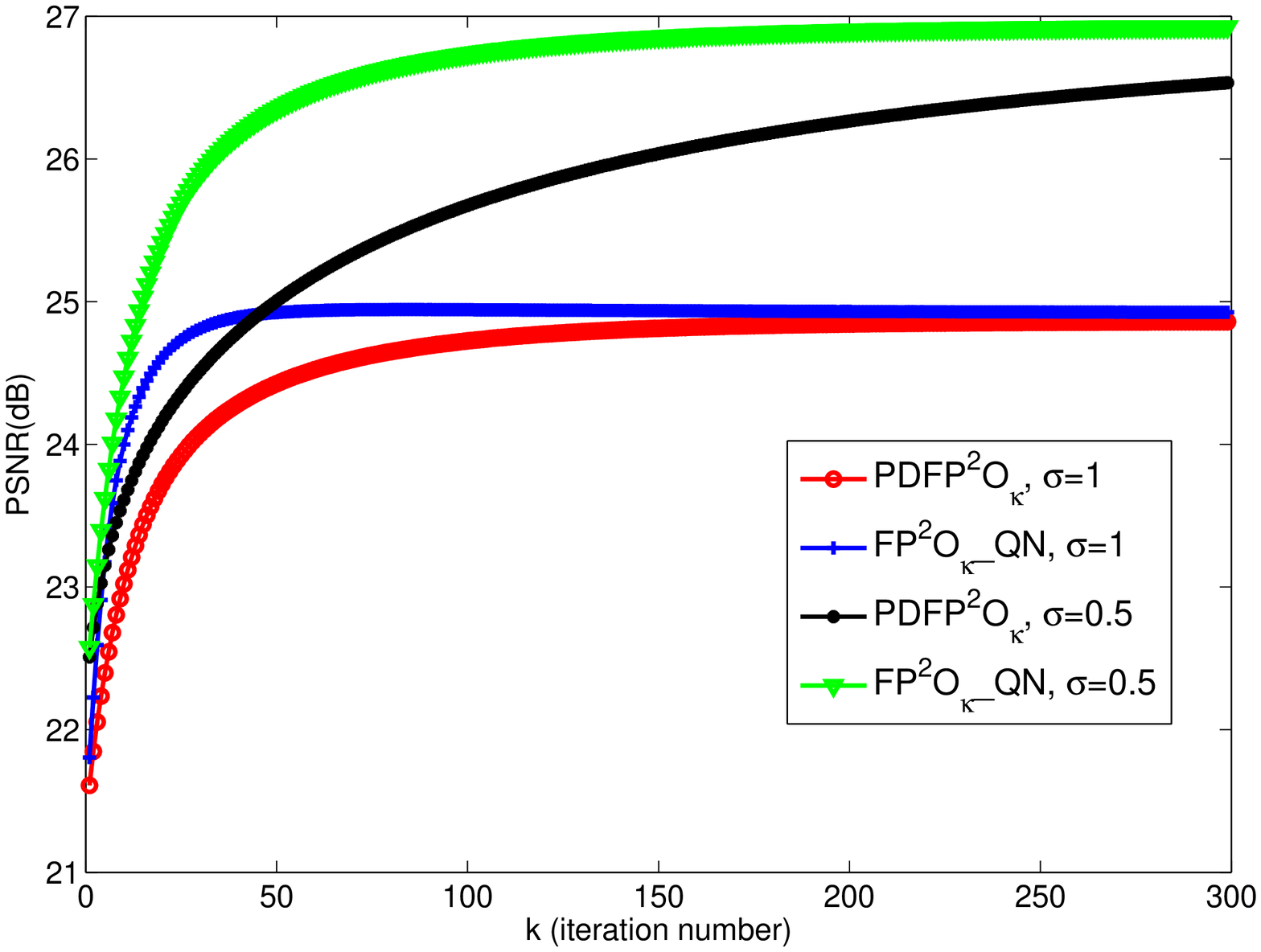}}
\caption{The evolution curves of PSNR (dB) for Pepper image with different blur kernels and noise levels. (a) Pepper image blurred by $8\times 8$ box average kernel; (b) Pepper image blurred by $6\times 6$ gaussian kernel.
}
\label{fig4.5}
\end{figure}

\begin{figure}
  \centering
  \subfigure[]{
    \label{fig4.6:subfig:a} 
    \includegraphics[width=2.5in,clip]{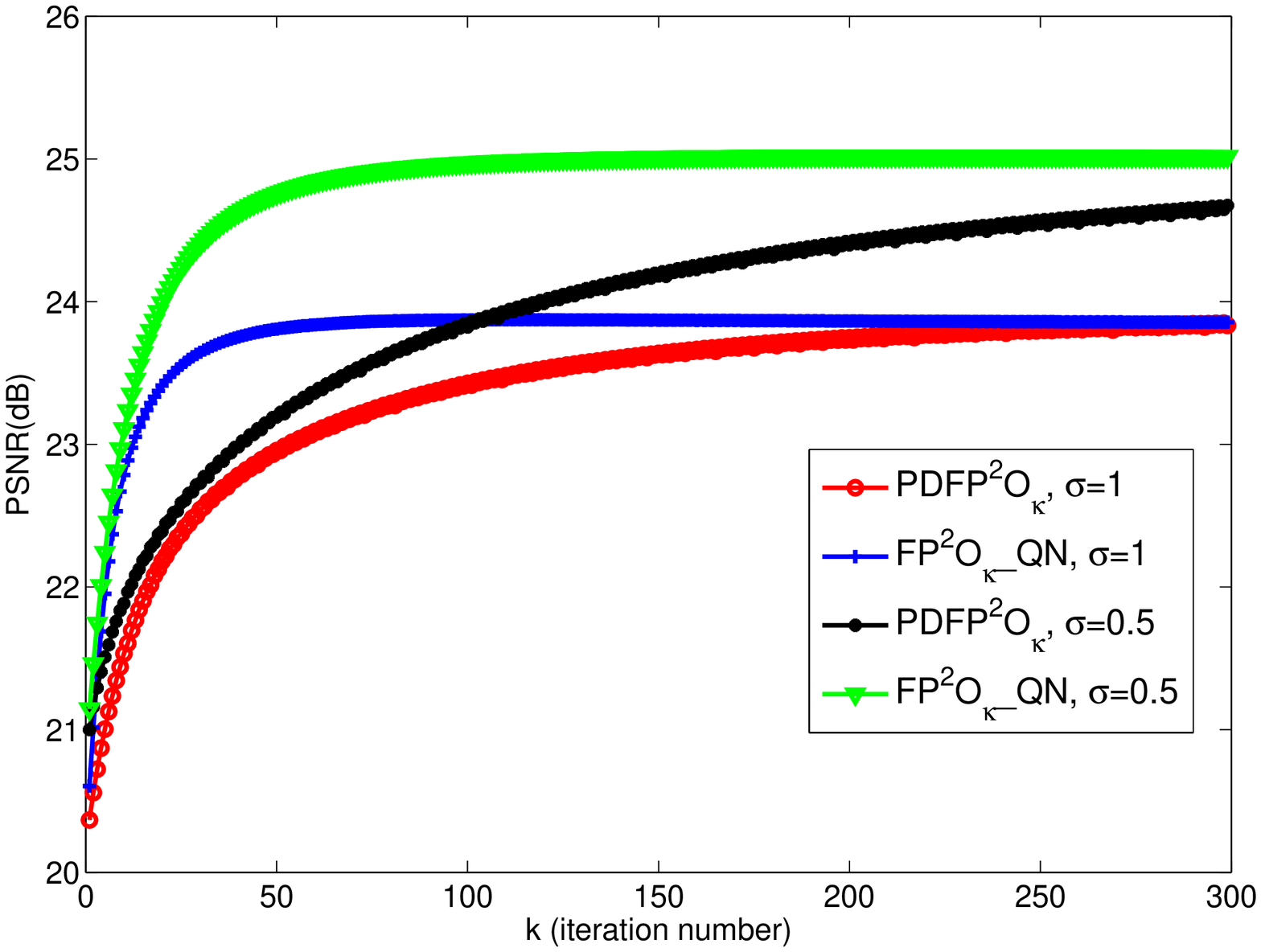}}
  \subfigure[]{
    \label{fig4.6:subfig:b} 
    \includegraphics[width=2.45in,clip]{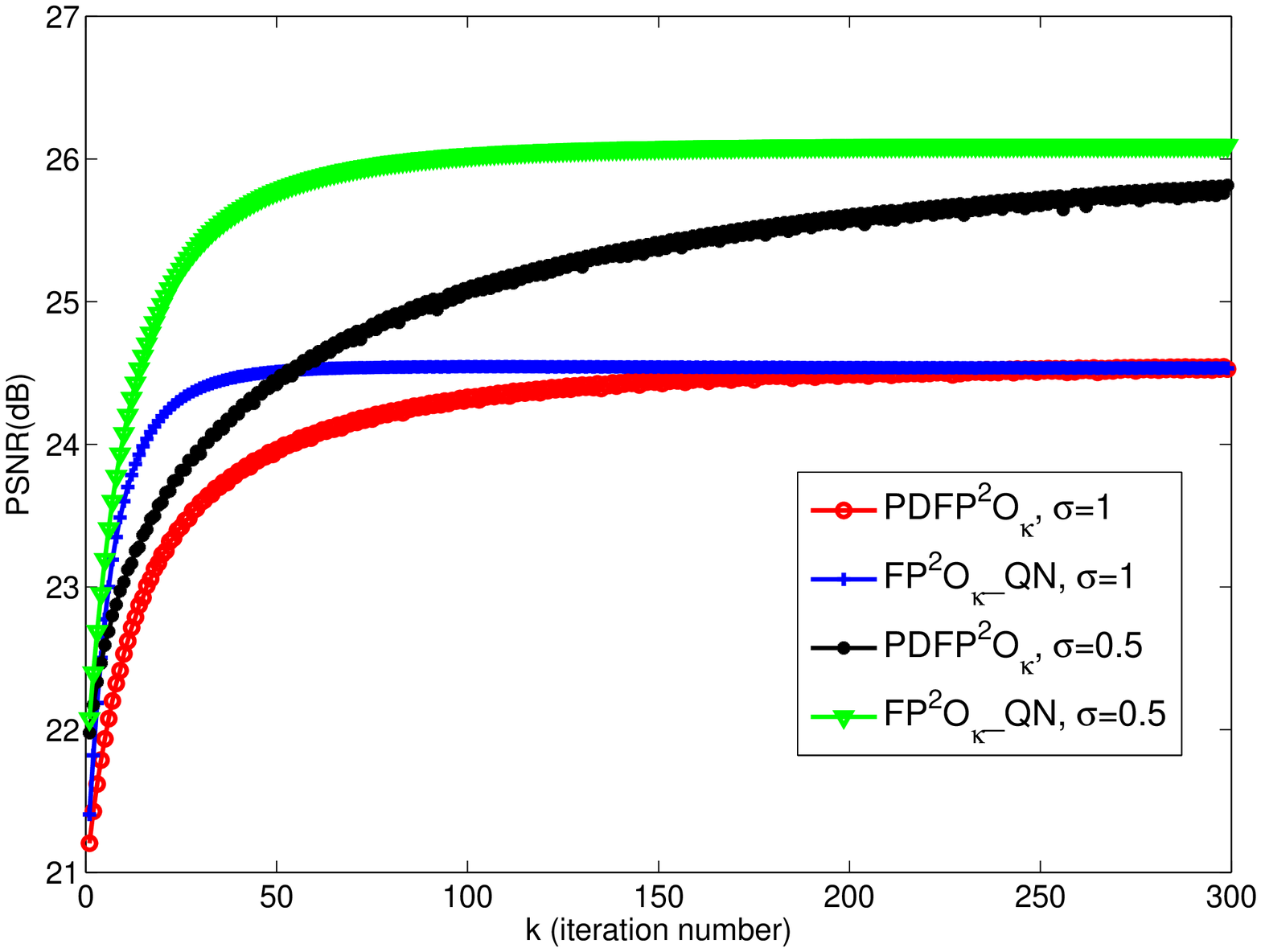}}
\caption{The evolution curves of PSNR (dB) for Cameraman image with different blur kernels and noise levels. (a) Cameraman image blurred by $8\times 8$ box average kernel; (b) Cameraman image blurred by $6\times 6$ gaussian kernel.
}
\label{fig4.6}
\end{figure}

In what follows, two images, Pepper (with size of $256\times256$) and Cameraman, are used for our test. Figures \ref{fig4.5}--\ref{fig4.6} show the evolution curves of PSNR (dB) running both fixed point algorithms for the two images. From the plots we observe that the PSNR values obtained by $\textrm{FP^{2}O_{\kappa}\_QN}$ increase much faster than those by $\textrm{PDFP^{2}O_{\kappa}}$. This is due to the quasi-Newton method included in $\textrm{FP^{2}O_{\kappa}\_QN}$ is more efficient than the gradient descent algorithm involved in $\textrm{PDFP^{2}O_{\kappa}}$.

\begin{figure}
  \centering
  \subfigure[]{
    \label{fig4.7:subfig:a} 
    \includegraphics[width=2.0in,clip]{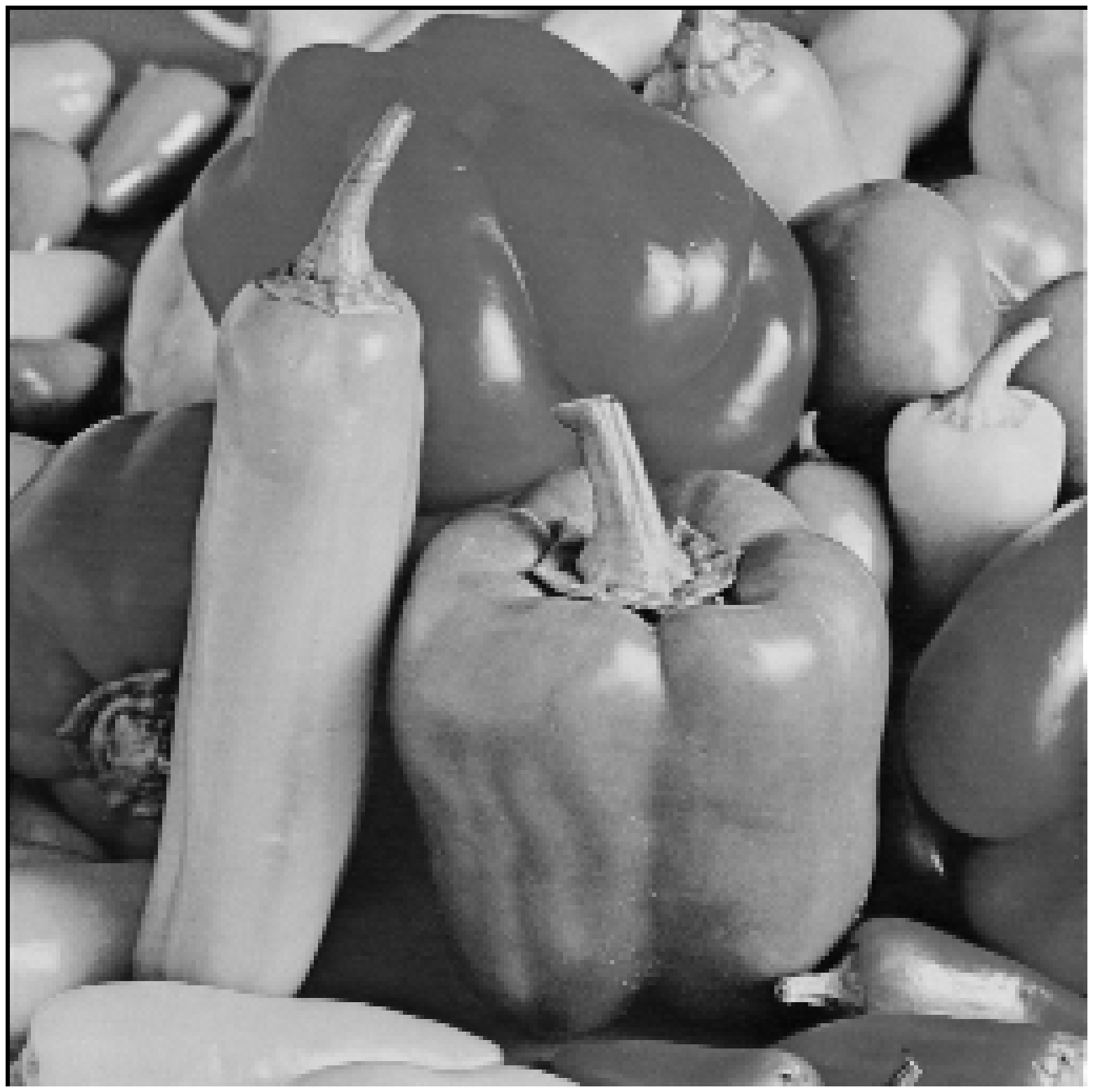}}
  \subfigure[]{
    \label{fig4.7:subfig:b} 
    \includegraphics[width=2.0in,clip]{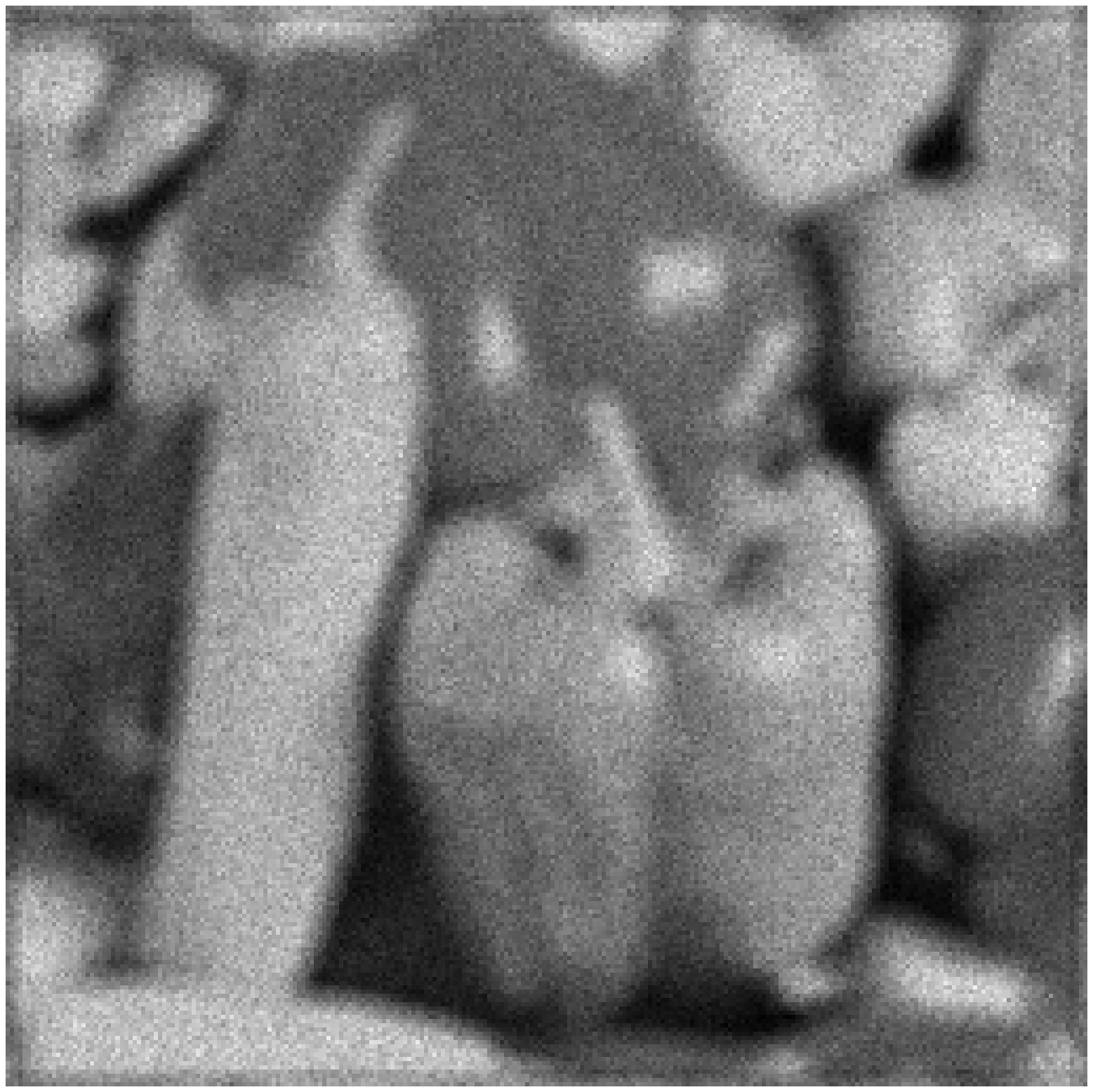}}
  \subfigure[]{
    \label{fig4.7:subfig:c} 
    \includegraphics[width=2.0in,clip]{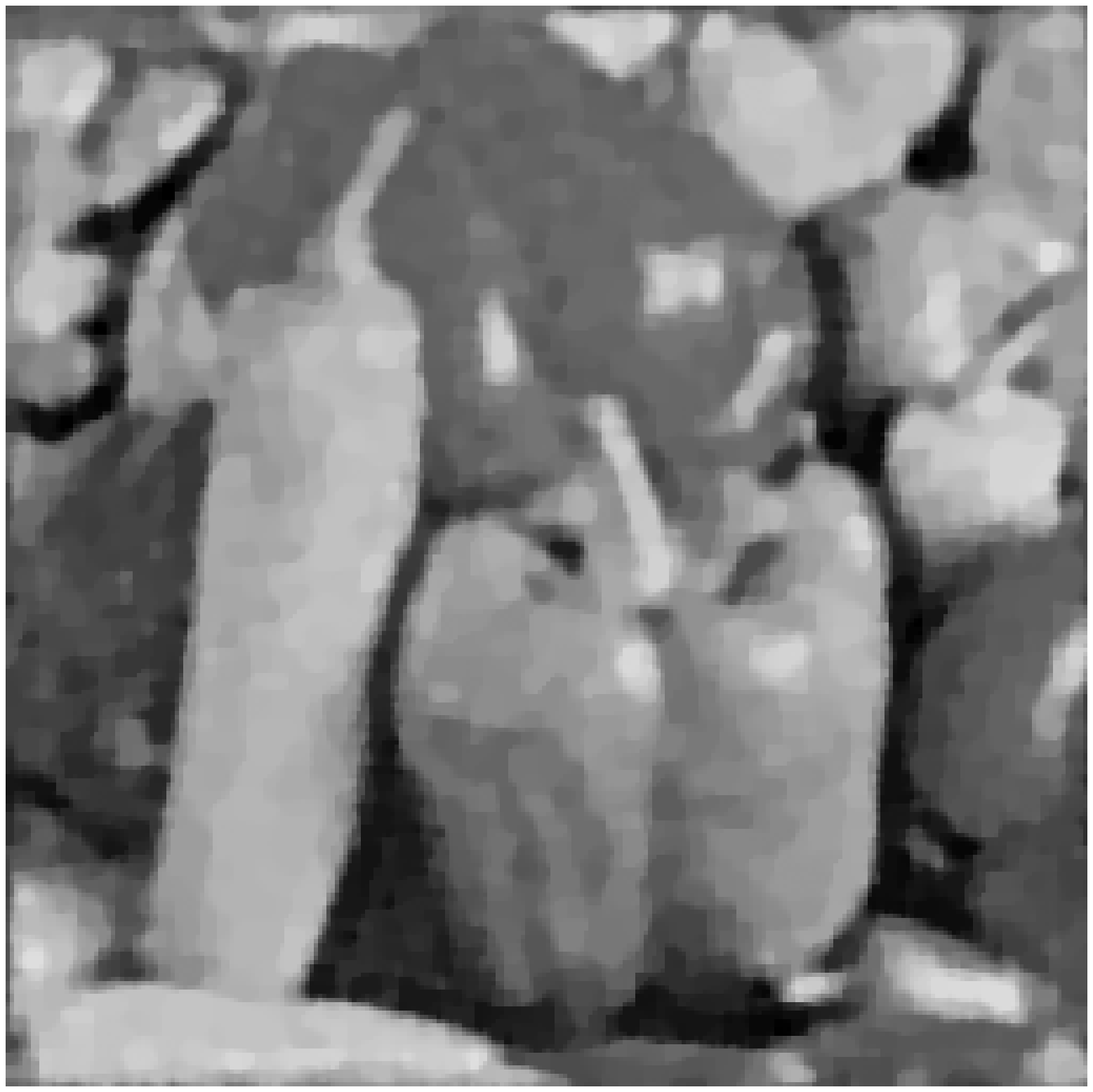}}
  \subfigure[]{
    \label{fig4.7:subfig:d} 
    \includegraphics[width=2.0in,clip]{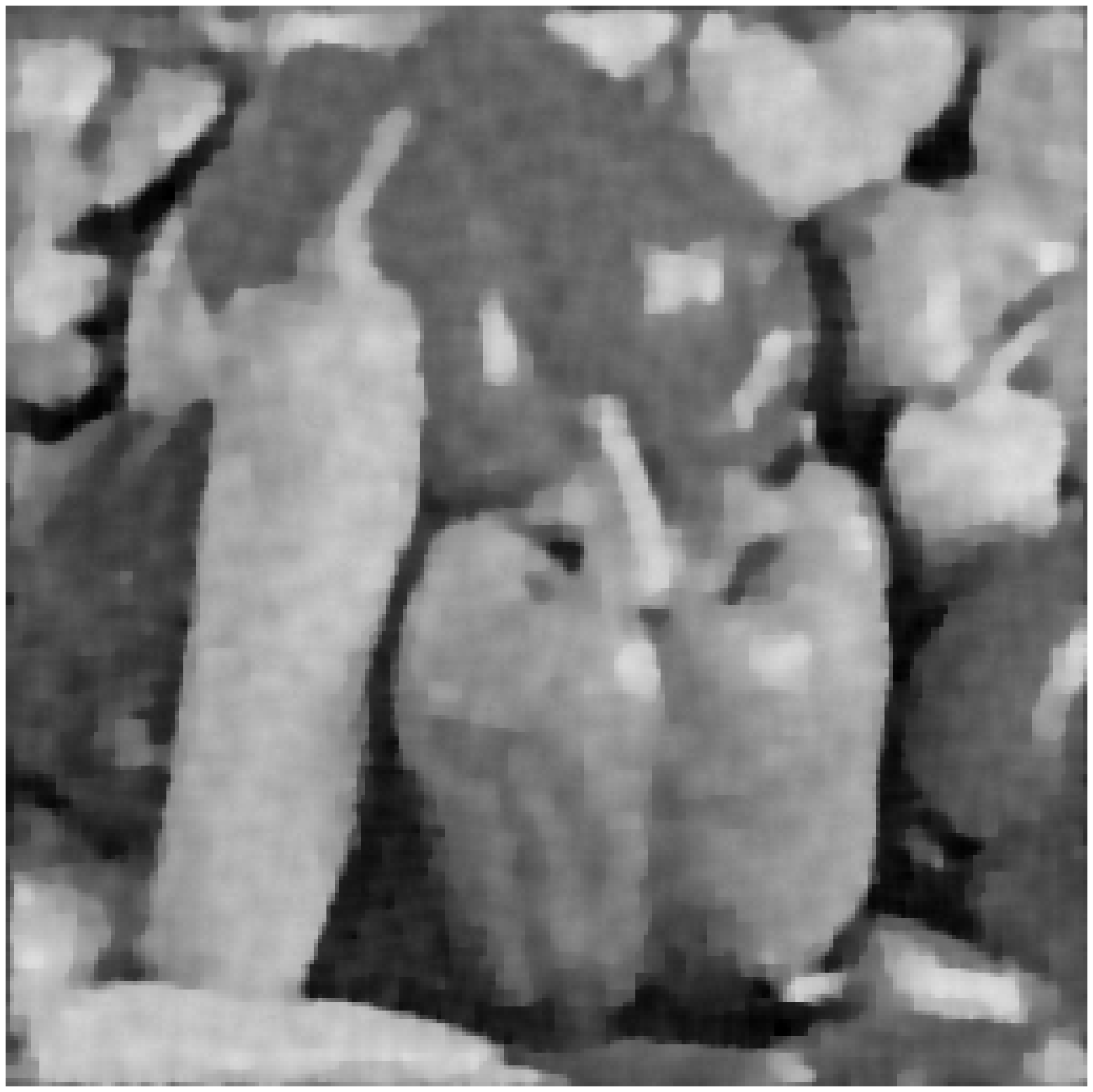}}
\caption{(a) The original Pepper image, (b) the blurry and noisy image, (c) the image restored by $\textrm{PDFP^{2}O_{\kappa}}$, PSNR=23.83, iter.=115, time=2.29, (d) the image restored by $\textrm{FP^{2}O_{\kappa}\_QN}$, PSNR=24.04, iter.=48, time=1.40.
}
\label{fig4.7}
\end{figure}

\begin{figure}
  \centering
  \subfigure[]{
    \label{fig4.8:subfig:a} 
    \includegraphics[width=2.0in,clip]{Cameraman.eps}}
  \subfigure[]{
    \label{fig4.8:subfig:b} 
    \includegraphics[width=2.0in,clip]{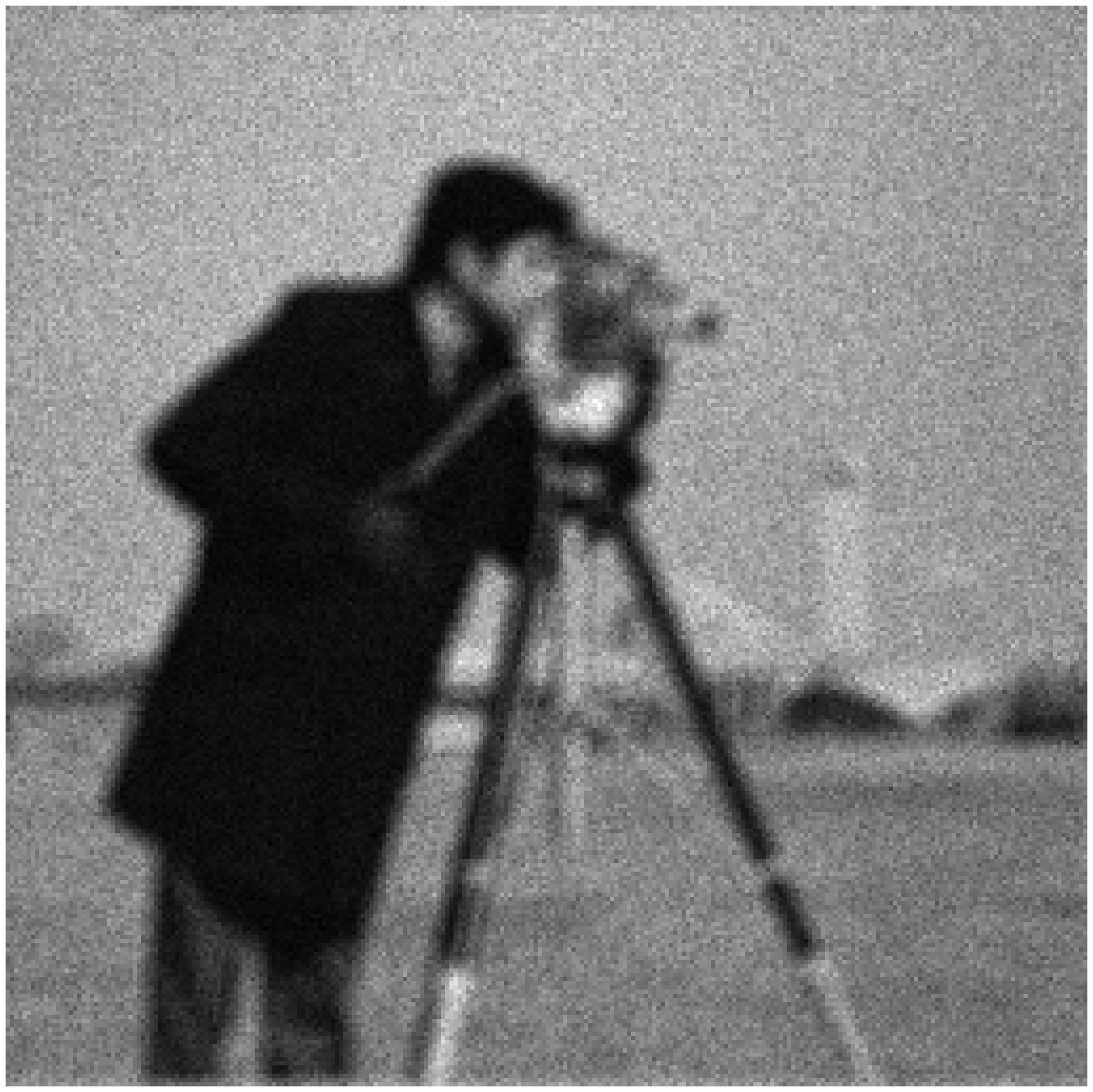}}
  \subfigure[]{
    \label{fig4.8:subfig:c} 
    \includegraphics[width=2.0in,clip]{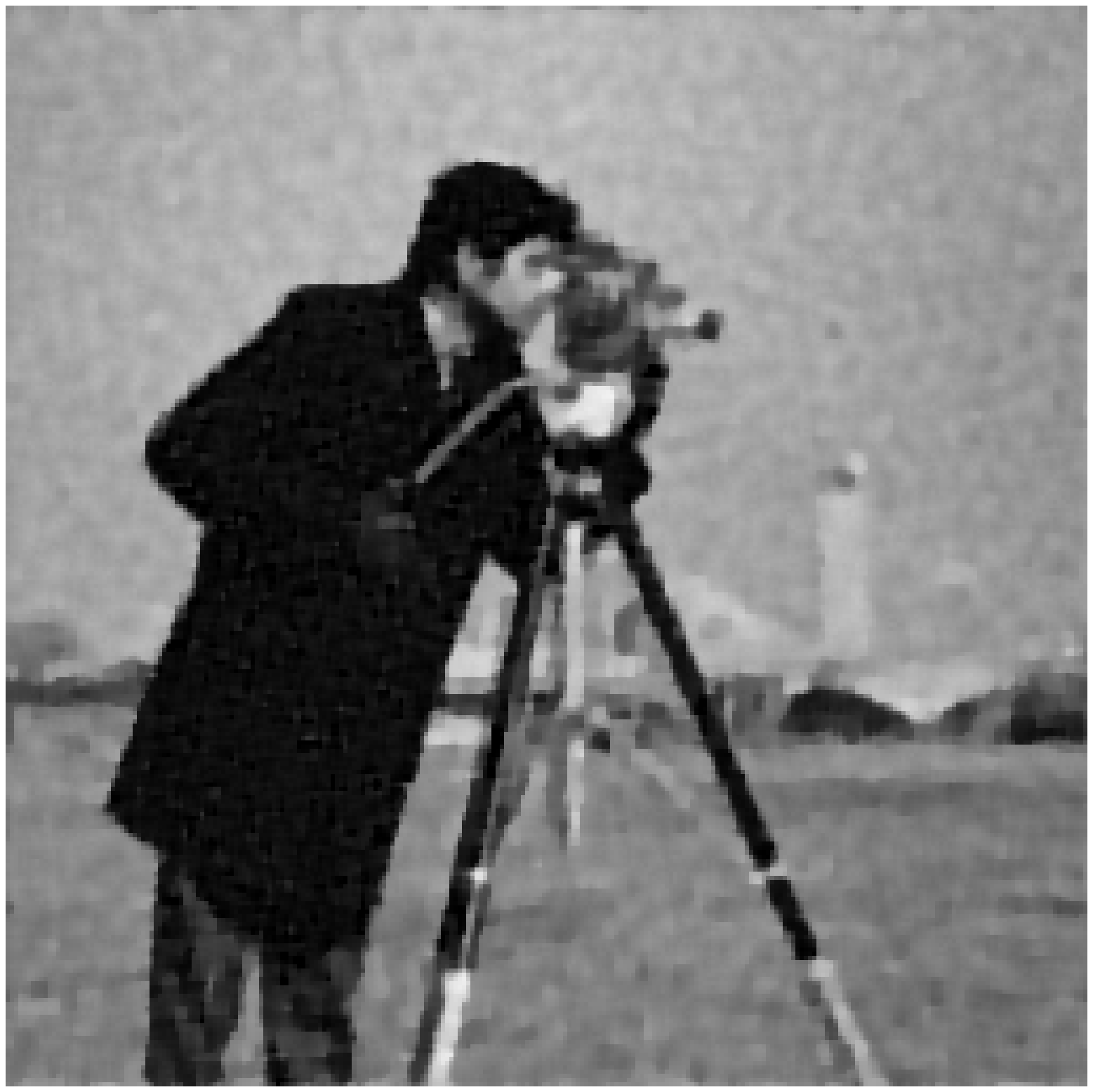}}
  \subfigure[]{
    \label{fig4.8:subfig:d} 
    \includegraphics[width=2.0in,clip]{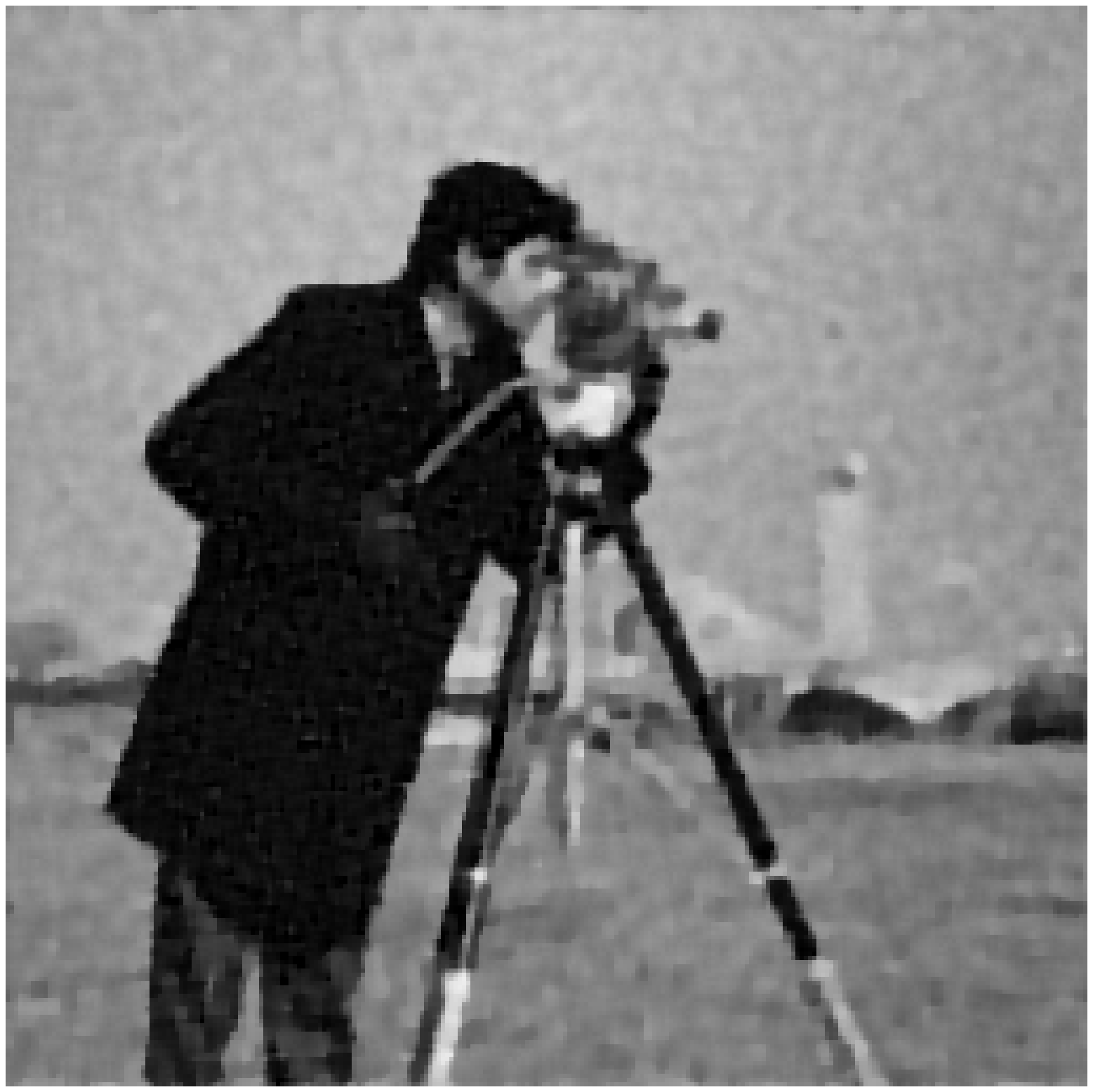}}
\caption{(a) The original Cameraman image, (b) the blurry and noisy image, (c) the image restored by $\textrm{PDFP^{2}O_{\kappa}}$, PSNR=24.36, iter.=108, time=2.24, (d) the image restored by $\textrm{FP^{2}O_{\kappa}\_QN}$, PSNR=24.49, iter.=45, time=1.14.
}
\label{fig4.8}
\end{figure}

Figure \ref{fig4.7} shows the deblurred results of Pepper image convoluted by $8\times 8$ box average kernel and contaminated by Rayleigh noise with $\sigma=1.0$. The corresponding PSNR values, the number of iterations, and the CPU time are also included. It is observed that $\textrm{FP^{2}O_{\kappa}\_QN}$ can obtain higher PSNR values with less iteration number and running time compared to  $\textrm{PDFP^{2}O_{\kappa}}$. The recovery results of Cameraman image blurred by $6\times 6$ gaussian kernel and corrupted by Rayleigh noise with $\sigma=1.0$ are also presented in Figure \ref{fig4.8}. We also observe that $\textrm{FP^{2}O_{\kappa}\_QN}$ is more efficient than $\textrm{PDFP^{2}O_{\kappa}}$, especially in the implementation time.

\section{Conclusion}\label{sec5}

In this article, we propose a fast fixed point algorithm based on the quasi-Newton method, abbreviated as $\textrm{FP^{2}O_{\kappa}\_QN}$, for solving the minimization problems with the general form of $\min_{u \in
\mathbb{R}^{N}}\left\{f_{1}(Bu)+f_{2}(u)\right\}$. The main distinction between $\textrm{FP^{2}O_{\kappa}\_QN}$ and previous fixed point algorithms lies in that the quasi-Newton method, rather than the gradient descent algorithm, is included in the algorithm framework. The proposed algorithm framework is applied to solve TV-based image restoration problem. Numerical experiments reported in this paper indicate that $\textrm{FP^{2}O_{\kappa}\_QN}$ outperform the recently proposed $\textrm{PDFP^{2}O_{\kappa}}$, especially in the implementation time.

\section{Acknowledgments}\label{sec6}
The research was supported in part by the National Natural Science Foundation of China under Grant 61271014.

\end{document}